\documentclass{article} 
\usepackage{iclr2025_conference,times}

\usepackage{amsmath,amsfonts,bm}

\def\eqref#1{equation~\ref{#1}}

\def\1{\bm{1}}

\DeclareMathAlphabet{\mathsfit}{\encodingdefault}{\sfdefault}{m}{sl}
\SetMathAlphabet{\mathsfit}{bold}{\encodingdefault}{\sfdefault}{bx}{n}

\usepackage{silence}
\ActivateWarningFilters[pdftoc]
\usepackage{hyperref}
\usepackage{url}
\usepackage{graphicx}
\usepackage{multirow}
\usepackage{amssymb}
\usepackage{caption}
\usepackage{booktabs}
\usepackage{colortbl}
\usepackage{adjustbox}

\title{Black Sheep in the Herd: Playing with \\ Spuriously Correlated Attributes for \\ Vision-Language Recognition}

\iclrfinalcopy
\author{Xinyu Tian\textsuperscript{\rm 1}, Shu Zou\textsuperscript{\rm 1}, Zhaoyuan Yang\textsuperscript{\rm 2}, Mengqi He\textsuperscript{\rm 1}, Jing Zhang\textsuperscript{\rm 1}\\ 
\textsuperscript{\rm 1}Australian National University, \quad \textsuperscript{\rm 2}GE Research
}

\newcommand\ChangeRT[1]{\noalign{\hrule height #1}}
\newcommand{\NA}{---}
\newcommand{\rotbox}[1]{\rotatebox{55}{#1}}

\begin{document}

\maketitle

\begin{abstract}
Few-shot adaptation for Vision-Language Models (VLMs) presents a dilemma: balancing in-distribution accuracy with out-of-distribution generalization. Recent research has utilized low-level concepts such as visual attributes to enhance generalization. However, this study reveals that VLMs overly rely on a small subset of attributes on decision-making, which co-occur with the category but are not inherently part of it, termed spuriously correlated attributes. This biased nature of VLMs results in poor generalization. To address this, 1) we first propose {\sc Spurious Attribute Probing} (\texttt{SAP}), identifying and filtering out these problematic attributes to significantly enhance the generalization of existing attribute-based methods; 2) We introduce {\sc Spurious Attribute Shielding} (\texttt{SAS}), a plug-and-play module that mitigates the influence of these attributes, seamlessly integrating into various Parameter-Efficient Fine-Tuning (PEFT) methods. In experiments, \texttt{SAP} and \texttt{SAS} significantly enhance accuracy on distribution shifts across 11 datasets and 3 generalization tasks while preserving downstream performance, establishing a new state-of-the-art benchmark. The code will be available \href{https://github.com/Liam-Tian/sas}{\textcolor[HTML]{d12d8a}{here}}.
\end{abstract}

\section*{1 \quad Introduction}
\label{sec:intro}
The emergence of large-scale pre-trained Vision-Language Models (VLMs)~\citep{radford2021learning, li2022blip} bridges the gap between images and texts. However, conventional fine-tuning of these models entails significant computational burdens, leading to Parameter-Efficient Fine-Tuning (PEFT), such as prompt tuning~\citep{MaPLe, Zhou_2022}, adapters~\citep{sung2022vl, gao2024clip} and LoRA~\citep{hu2021lora, dettmers2024qlora}. With PEFT, requiring approximately 1\% of model parameters, one may adeptly adapt to downstream tasks, achieving comparable or even superior performance to full fine-tuning~\citep{liu2021p}. Yet, recent studies have revealed that in few-shot scenarios where observed samples are limited, PEFT struggles to generalize to out-of-distribution datasets and may compromise the VLMs' strong zero-shot capability~\citep{zhou2022conditional, yao2023visual, bulat2023lasp}. This creates a trade-off where individuals aim for strong performance on downstream tasks while endeavoring to maintain the ability of VLMs to handle distribution shifts.

In response to the above-mentioned challenges, various strategies have been proposed such as category conditioning~\citep{zhou2022conditional, yao2023tcp}, prompt regularization~\citep{yao2023visual, bulat2023lasp, khattak2023self} and training-free adaptation~\citep{udandarao2022sus, zhang2021tip}. Recently, it has been discovered that incorporating descriptors, also known as visual attributes during training can significantly improve the accuracy of adapted modules on out-of-distribution datasets~\citep{zhang2024concept, liao2023descriptor, tian2023argue, ma2023attrseg, liu2024multi}. The motivation behind these works is that attributes, as lower-level concepts, are more likely to establish connections to unseen categories compared to the high-level names~\citep{zhang2024concept, tian2023argue}. These methods can be divided into two types: one involves generating visual attributes for the target category using Large Language Models (LLMs)~\citep{tian2023argue, ma2023attrseg, liu2024multi}, while the other entails searching for optimal attributes from a pre-defined vocabulary that maximizes semantic similarity~\citep{zhang2024concept} or training accuracy~\citep{liao2023descriptor}. Yet, a commonality among them is their reliance on the set of generated attributes, \ie, the attribute pool.

While promising, the limitations of this line of work have been underexplored. Initial suspicions emerge from \citet{roth2023waffling}, which find that in certain cases, replacing attributes with random sequences does not lead to a notable performance decline. Subsequently, \citet{an2023more} discover that VLMs sometimes disregard the presence of attributes, leading to minimal gains. This prompts us to inquire: \textit{Are attributes truly dependable? If so, then whence do these failure cases arise?}

To tackle the aforementioned inquiries, we conduct a manual examination of the attribute pool generated by existing methods. We stumble upon an often overlooked fact: while most attributes accurately depict the intrinsic characteristics of the target category, there exists a small subset of attributes that co-occur with the category but are not part of it, leading to strong spurious correlations. For instance, when querying LLM with ${\rm what \ does \ a \ mountain \ bike \ look \ like}$ we receive attributes like ${\rm wheels}, {\rm handle}$, and ${\rm basket}$, yet unexpected attributes such as ${\rm trees}$ and ${\rm road}$ also emerge. This phenomenon is also observed in vocabulary-based methods, where attributes are chosen based on in-distribution samples. Inspired by \citet{singla2021salient}, we refer to the former as \textit{core attributes} and the latter as \textit{spuriously correlated attributes}\footnote{We refer spuriously correlated attributes to spurious attributes for brevity.}.

\begin{figure}[t!]
 \begin{minipage}[c]{0.67\textwidth}
    \includegraphics[width=\textwidth]{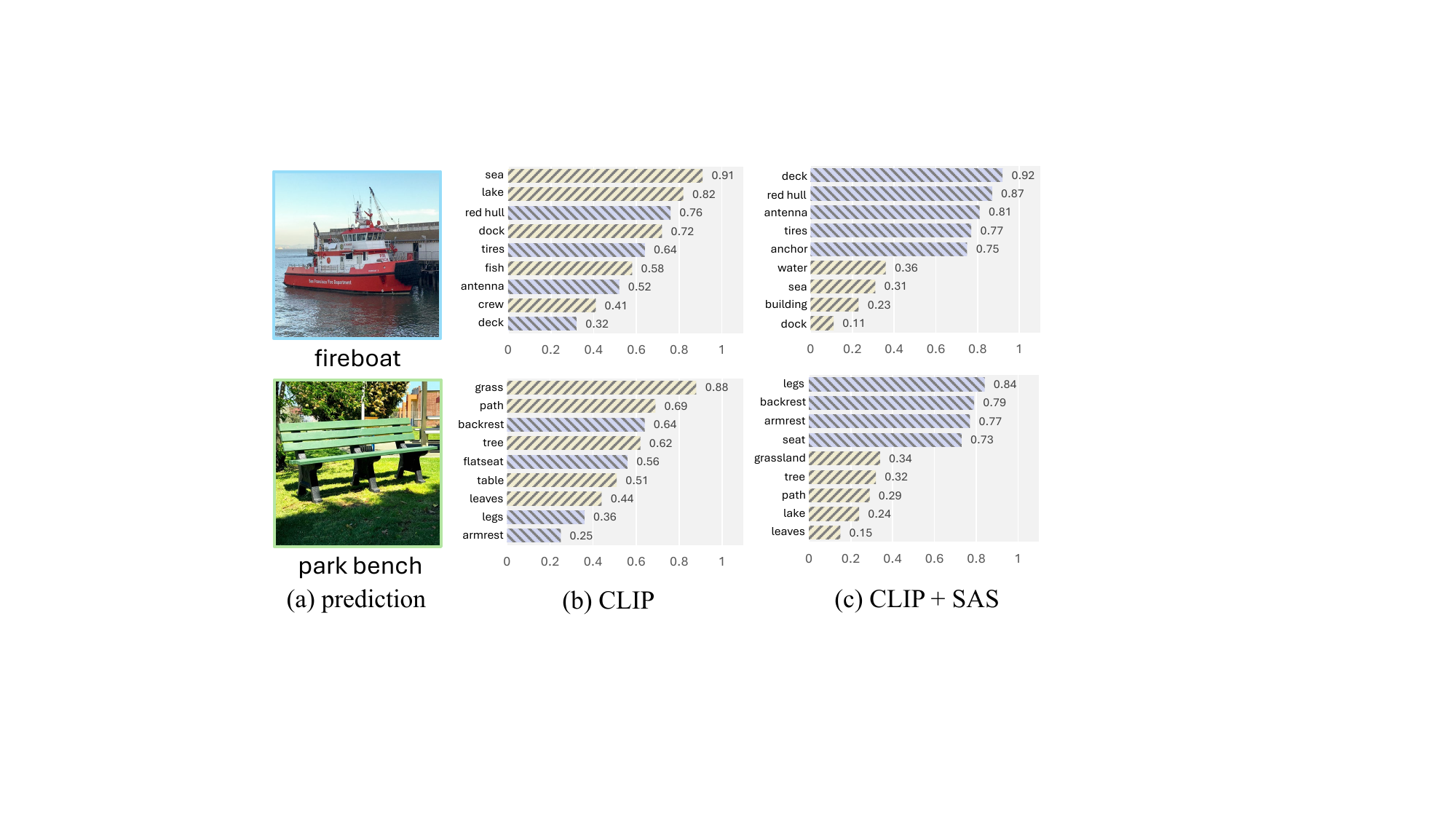}
  \end{minipage}
  \hfill
\begin{minipage}[c]{0.30\textwidth}
    \caption{\textbf{The phenomenon of Black Sheep in the Herd}. We rank attribute weights on VLM predictions using CBMs, with \textcolor[HTML]{DAC54F}{yellow} and \textcolor[HTML]{A4AEE7}{purple} bars to denote spurious and core attributes respectively. In \textbf{(b)}, we observe that for vanilla VLMs, 2 out of the top-3 are spurious attributes, heavily influencing decisions. In \textbf{(c)}, \texttt{SAS} mitigates this by suppressing the influence of spurious attributes.}
  \label{fig:intro}
  \vspace{-2mm}
  \end{minipage}
  \vspace{-6mm}
\end{figure}
Building upon the motivation to enhance generalization, a natural idea is to strive for a ``pure" attribute pool that accurately reflects the true characteristics of the category. Therefore, we conduct a simple experimental study based on existing methods~\citep{zhang2024concept, tian2023argue} where we manually identify attributes that might lead to spurious correlations with the target category and remove them. Despite the small proportion of these attributes ($<7\%$),  we observe a significant improvement in out-of-distribution generalization accuracy. To gain deeper insights into how spurious attributes affect VLMs, we employ concept bottleneck models (CBMs)~\citep{koh2020concept, yang2023language}, a well-established method for interpreting and ranking attribute weights in decision-making processes. Our analysis reveals that despite their small representation in the overall pool, spurious attributes exert a significant influence, particularly among the top-3 attributes on decision-making as depicted in Fig.~\ref{fig:intro}. We term this phenomenon \textit{Black Sheep in the Herd} since 1) spurious attributes act as the ``black sheep" within the pool, constituting a small fraction; 2) nevertheless, this small fraction significantly impacts the generalization ability of VLMs.

We could utilize the aforementioned manual inspection to assist existing attribute-based methods. However, this remains a mere fancy dream since manually identifying spurious attributes in the pool is prohibitively expensive. This motivates us to devise a new method for generating a pure pool, one that contains only those core attributes belonging to the category. Hence, we propose {\sc Spurious Attribute Probing} (\texttt{SAP}), an approach to derive an attribute pool where core and spurious attributes are clearly separated. \texttt{SAP} integrates Multi-modal Large Language Models (MLLMs) and Concept Bottleneck Models (CBMs) to tackle this challenge. Leveraging MLLMs, \texttt{SAP} initially distinguishes core attributes from non-core counterparts, and then CBMs prioritize the latter by selecting those with a significant impact on model decisions as spurious attributes. \texttt{SAP} complements existing attribute-based methods and, to the best of our knowledge, is the first approach to identifying spurious attributes in open-world settings without explicit human supervision.

Despite the effectiveness of \texttt{SAP}, it faces limitations: 
It may prevent the presence of spurious attributes in the language branch, yet it cannot stop the model from learning spurious features. This extends the scope beyond attribute-based methods, and will result in poor generalization across PEFT family. Therefore, we propose {\sc Spurious Attribute Shielding} (\texttt{SAS}), a module to mitigate the influence of spurious features which can be seamlessly integrated into arbitrary PEFT methods. Specifically, \texttt{SAS} introduces a subsidiary task by creating a set of pseudo categories defined by spurious attributes alongside the real ones, allowing VLMs to distinguish between them. For instance, if ${\rm streetlight}$ is considered a spurious attribute for the target category ${\rm vehicle}$, we establish a separate pseudo category exclusively for ${\rm streetlight}$ and discern between the two, thus decreasing the dependency on ${\rm streetlight}$ for identifying ${\rm vehicle}$. The experiments show that by combining \texttt{SAS} into existing PEFT approaches, the accuracy under distribution shifts is significantly improved, reaching a new state of the art.

In summary, our main contributions are as follows:
\begin{itemize}
    \item Despite the promise of visual attributes in various applications, we discover a group of \textit{black sheep}, \ie, spurious attributes, on which VLMs inherently heavily rely, thereby leading to poor generalization and robustness.
    \item We introduce {\sc Spurious Attribute Probing} (\texttt{SAP}), aiming to identify and eliminate these problematic attributes, thereby substantially improving the generalization of current attribute-based methods.
    \item We present {\sc Spurious Attribute Shielding} (\texttt{SAS}), a plug-and-play module seamlessly integrating into various PEFT methods to mitigate the influence of spurious attributes on predictions.
\end{itemize}
\section*{2 \quad Related Work}
\label{sec:related}
{\textbf{Vision-Language Models.} Recently, it has been discovered that associating text and images for pre-training, instead of using images alone, enables powerful zero-shot capability, leading to VLMs. Initially, simple dual-tower structures are employed, where the representations of the two modalities are modeled by separate encoders and connected via contrastive learning, \ie, CLIP~\citep{radford2021learning}. Subsequently, more works have been built upon this foundation. For instance, \citet{li2022blip} bridge two encoders by fusion for better cross-modality interactions, \citet{li2023scaling} employ masked image modeling to achieve a trade-off between accuracy and training time, and \citet{li2022grounded} incorporate visual detection and grounding in pre-training for object-level reasoning. For more information, we refer to \citet{zhang2024vision} for a detailed survey of recent VLMs.}

\textbf{Parameter-Efficient Fine-Tuning.} As pre-trained models grow larger, traditional fine-tuning demands significant resources, leading to PEFT~\citep{hu2021lora, liu2021p, lester2021power, houlsby2019parameter}. However, PEFT is a double-edged sword: while it adeptly adjusts to downstream tasks, it also brings poor generalization to the open world, inspiring various current remedies. For instance, category conditioning~\citep{zhou2022conditional, yao2023tcp} infuses category-aware knowledge for discriminative and generalizable learning, prompt regularization~\citep{yao2023visual, bulat2023lasp} confines learnable prompts to corresponding textual features, and training-free adaptation~\citep{udandarao2022sus, zhang2021tip} eschews gradient-based optimization to prevent overfitting. Recently, another line of work leveraging visual attributes has shown promising results, achieving state-of-the-art performance in various generalization tasks.

\textbf{Visual Attributes for Recognition.} The initial exploration of attributes for recognition begins in zero-shot settings~\citep{menon2022visual, pratt2023does}, where individuals utilize attributes generated by LLMs to offer more expressive and accurate descriptions. Subsequently, it is observed that training VLMs to grasp fundamental concepts like visual attributes aids in generalizing to unseen data, prompting a surge in attribute-based methods. For instance, \citet{tian2023argue} appends attributes to category names, \citet{liao2023descriptor} initializes learnable tokens as attribute embeddings, and \citet{ma2023attrseg} adopts a more aggressive approach by replacing category names entirely with attributes. {Additionally, \citet{wei2019adversarial} utilize adversarial training to learn attribute-object composition, while \citet{huang2024open} and \citet{wang2015multiple} improve the model's fine-grained understanding by building multi-granularity and hierarchical attributes.} Nonetheless, recent studies have noted a decline in attribute effectiveness in certain scenarios~\citep{an2023more}, sometimes reducing to a mere ensembling effect~\citep{roth2023waffling}. This paper delves into the issue, attributing it to spurious attributes, and proposes two plug-and-play approaches to complement existing methods.

\textbf{Spurious Attribute Identification.} Spurious attributes arise from model debiasing~\citep{seth2023dear, chuang2023debiasing, berg2022prompt}, defined as \textit{those likely to co-occur with the object but not part of it}~\citep{singla2021salient}. Although a well-known term, it remains underexplored due to the difficulty in identification. The initial endeavor by \citet{singla2021salient} involves manually labeling to identify spurious attributes. Similarly, \citet{wong2021leveraging} integrates human supervision with sparse linear layers to mitigate labor expenses. Others identify spurious attributes by analyzing their properties. For instance, \citet{wu2023discover} observe that spurious attributes exhibit instability across data environments and introduce concept sensitivity for identification. Conversely, \citet{teotia2022finding} train an attribute probing network to predict spurious attributes. Recently, the work most akin to ours, \citet{adila2023zero}, utilize LLMs to derive harmful insight representations by comparing differences between concepts. However, the inference complexity of this method escalates exponentially with the number of concepts, restricting its application to small-scale datasets. In contrast, our proposed \texttt{SAP} 1) necessitates neither human labeling nor a training process, rendering it extremely cost-effective; and 2) is scalable to any large-scale dataset, \eg, ImageNet.

\textbf{Spurious Correlation Mitigation.} Current spurious mitigation methods can be mainly categorized into two types. The first assumes that spurious attributes within the dataset are either unknown or complex, employing various proxies to mitigate spurious correlations. For instance, \citet{xu2020adversarial, yao2022improving, han2022umix} introduce augmentation via domain mix-up to learn invariant features, while \citet{li2022discover, zhang2022correct, utama2020towards} advocate for instance reweighting to emphasize hard samples. Others calibrate biased representation through contrastive learning~\citep{you2024calibrating, zhang2022contrastive}. The second type explicitly assumes that spurious correlations arise from known attributes~\citep{chuang2023debiasing, berg2022prompt}. For instance, \citet{wu2023discover} balance training data by swapping spurious concepts among categories, whereas \citet{adila2023zero} calibrate embeddings by removing spurious representations. In contrast, \texttt{SAS} belongs to the latter, where the attribute prior is known, and thanks to the effectiveness of \texttt{SAP}, it may accurately mitigate spurious correlations caused by identified spurious attributes. For further details, a quantitative comparison between \texttt{SAS} and related works is provided in Supp. Mat. \hyperref[subsec:B7]{B}.
\section*{3 \quad Method}
\label{sec:method}
\subsection*{3.1 \quad Problem Setup}
\label{subsec:setup}
We assume the training set of pairs $\mathcal{D} = \{(x, c)\}$, where $x \in \mathcal{X}$ and $c \in \mathcal{C}$ represent the image and ground truth label, respectively. The attribute-based methods aim to construct a category-wise prompt $t_{c} = f(c, \mathcal{P})$ such that the conditional distribution of the prediction $y$ given $x$ is modeled as
\begin{equation}
\label{eq:probability}
     P(y \vert x) = \frac{\exp(s(\phi_{I}(x), \phi_{L}(t_{y})) / \tau)}{\sum\limits_{c \in \mathcal{C}}\exp(s(\phi_{I}(x), \phi_{L}(t_{c})) / \tau)},
\end{equation}
where $\phi_{I}$ and $\phi_{L}$ represent the vision and language encoder, respectively, $s(\cdot, \cdot)$ indicates the similarity function, and $\tau$ is the temperature scaler. $\mathcal{P} = \{\mathcal{A}_{c}\mid{c\in \mathcal{C}}\}$ is an attribute pool generated by $\mathcal{A}_{c} = \mathcal{U}(\mathcal{H}(c))$ such that $\mathcal{A}_{c} = \{a_{c}^{1}, a_{c}^{2}, ..., a_{c}^{J}\}$. Depending on previous work, $\mathcal{U}$ could be a LLM, thus $\mathcal{H}(c)$ is a set of LLM prompts incorporating the category name of $c$. $\mathcal{U}$ could also be a large vocabulary, such that $\mathcal{H}(c)$ becomes a key to search for the semantically related attributes. Therefore, $f(\cdot, \cdot)$ is denoted as a concatenation function to integrate the category name and corresponding attributes together. The optimization objective is typically a cross-entropy loss $\mathcal{L}_{ce}$.

It's important to mention that in this work, we refrain from specifying particular learnable parameters; they could encompass learnable prompts, adapters, or LoRA. Our goal is to ensure the versatility of our plug-and-play method across various PEFT approaches.
\subsection*{3.2 \quad Motivation}
\label{subsec:motivation}
We present the motivation of this work by revealing an overlooked fact: the attribute pool in current methods are not purely aligned with the intrinsic semantics of categories. Some attributes are spurious, co-occurring with categories but not inherently linked to them. To investigate the impact of these ``black sheep", we conduct a simple experimental study. Specifically, we manually traverse the attribute pool $\mathcal{P}$ across various methods and identify spurious attributes within. We use a conventional method following \citet{singla2021salient} with a simplistic version. Given the category $c$, we randomly sample 5 images from the shots and visualize the heatmap. For specific attribute $a_{c}^{k}$, we determine whether it is a part of the main object, or separate objects in the background based on the sampled images with the heatmap activations. 

Upon identifying these spurious attributes, we remove them from the pool and compare the changes in their generalization capability before and after elimination. The experiment is evaluated on base-to-new generalization, following the outlined settings in Section~\hyperref[sec:experiment]{4}. As baselines, we select CPL~\citep{zhang2024concept} and ArGue~\citep{tian2023argue}, representing vocabulary-based and LLM-assisted methods, respectively. Additionally, we consider a variant, ArGue*, where we modify the LLM prompts to reduce the likelihood of spurious attribute occurrence. For instance, we append an additional instruction ${\rm focus \ on \ mountain \ bike \ itself}$ to the original prompt. Further details are provided in Supp. Mat. \hyperref[subsec:A1]{A}.

\textbf{Removing spurious attributes significantly enhances generalization.} 
While most attributes contribute positively to generalization, spurious attributes are exceptions to this trend. Removing these exceptions leads to a notable increase in accuracy on the new category on average ($65.30\% \rightarrow 67.66\%$ for CPL and $66.07\% \rightarrow 67.69\%$ for ArGue), without compromising accuracy on the base category. This phenomenon is aptly described as Black Sheep in the Herd since 1) spurious attributes constitute only a small portion of the pool ($<7\%$); 2) yet this small portion significantly impacts the generalization ability of VLMs.

\begin{table*}[t]
\footnotesize
\centering
\setlength{\tabcolsep}{0.8mm}
\renewcommand{\arraystretch}{1.1}
\begin{tabular}{lccccccccccccccc}
\ChangeRT{1.2pt}
\multicolumn{1}{c}{\multirow{2}{*}{Method}} & \multicolumn{3}{c}{FGVCAircraft} & \multicolumn{3}{c}{SUN397} & \multicolumn{3}{c}{Flowers102} & \multicolumn{3}{c}{DTD} & \multicolumn{3}{c}{Average} \\ \cline{2-16} 
\multicolumn{1}{c}{}                        & Base      & New      & SR      & Base      & New       & SR       & Base      & New      & SR      & Base   & New    & SR    & Base     & New     & SR     \\ \hline
CPL                                         & 42.27     & 38.85    & 5.43    & 81.88     & 79.65     & 6.61     & 98.07     & 80.43    & 5.71    & 80.92  & 62.27  & 5.13  & 75.79    & 65.30   & 5.72   \\
CPL - SA                                    & \textbf{42.62}     & 41.33    &  \NA       & 82.14     & \textbf{82.36}     &    \NA      & 98.35     & \textbf{82.16}    &    \NA     & \textbf{81.62}  & 64.77  &     \NA  & \textbf{76.18}    & 67.66   &    \NA    \\ \hline
ArGue                                       & 41.29     & 38.80    & 5.13    & 81.89     & 80.48     & 6.45     & 98.62     & 77.96    & 6.69    & 80.33  & 67.03  & 5.97  & 75.53    & 66.07   & 6.06   \\
ArGue*                                      & 41.34     & 39.34    & 4.86    & 81.97     & 80.62     & 6.11     & 98.58     & 78.11    & 6.44    & 80.41  & 67.26  & 5.62  & 75.58    & 66.33   & 5.76   \\
ArGue - SA                                  & 41.55     & \textbf{41.60}    &  \NA       & \textbf{82.33}     & 81.94     &    \NA      & \textbf{98.73}     & 78.75    &     \NA    & 80.79  & \textbf{68.47}  &   \NA    & 75.85    & \textbf{67.69}   &    \NA   \\ \ChangeRT{1.2pt}
\end{tabular}
\caption{\textbf{The results on base-to-new generalization before and after removing spurious attributes (SA) from the pool.} We report accuracy on base and new categories, and spurious rate (SR), which refers to the proportion of spurious attributes to the entire pool.}
\label{tab:preliminary}
\vspace{-8mm}
\end{table*}

\textbf{VLMs heavily rely on spurious attributes for predictions.} To deepen our understanding of this phenomenon, we use concept bottleneck models (CBMs)~\citep{koh2020concept} to determine attribute weights in model decision-making. In Fig.~\ref{fig:intro}, attributes are ranked by weight from high to low. Among the top-3 attributes influencing predictions, spurious attributes occupy two positions. For instance, in predicting ${\rm fireboat}$, VLMs heavily rely on ${\rm sea}$ and ${\rm lake}$ as crucial concepts, while for ${\rm park \ bench}$, ${\rm grass}$ and ${\rm path}$ are primary indicators. This implies that 1) VLMs may exhibit insensitivity to the presence of core attributes; 2) directly adapting to downstream tasks may heavily rely on spurious features for predictions. In fact, this also aligns with findings from concurrent work~\citep{wang2024clips}, which indicates that VLMs are more susceptible to spurious features compared to unimodal architectures.

In addition to the above observations, this prompts us to consider several questions.

\textbf{Where do spurious attributes come from?} In the case of LLMs, this phenomenon may be attributed to statistical bias in their large-scale training data. In practical scenarios, when describing a complex object, there may be a tendency to focus more on its accompanying scenes and associated elements rather than its core components. However, these accompanying elements may not contribute to a VLM's generalizable understanding of a specific category. Conversely, regarding vocabulary-based methods, their attribute selection heavily relies on in-distribution samples, and this preference for attributes may be also detrimental to generalization.

\textbf{Is there a better way to identify spurious attributes?} 
While manually purifying the attribute pool may enhance generalization, it faces two primary challenges: 1) scalability issues as the dataset size grows, and 2) it is a simple solution lacking quantitative assessment of the spurious correlation of each attribute, potentially leading to false positives, \ie, attributes that co-occur with the category merely by chance. Hence, we also experiment with an LLM-assisted variant called ArGue*, which adjusts the LLM prompts to encourage a stronger focus on the category itself. However, as demonstrated empirically in Table~\ref{tab:preliminary}, the reduction in the spurious rate is modest ($6.06\% \rightarrow 5.76\%$), resulting in only marginal gains ($66.07\% \rightarrow 66.33\%$).
\subsection*{3.3 \quad Spurious Attribute Probing}
\label{subsec:sap}
Motivated by the above considerations, we introduce {\sc Spurious Attribute Probing} (\texttt{SAP}), an approach to creating a comprehensive attribute pool where spurious and core attributes are distinctly separated. Initially, \texttt{SAP} utilizes Multi-modal Large Language Models (MLLMs) to differentiate attributes belonging to the target category, distinguishing core attributes from non-core counterparts. To determine if the coexistence of the latter with the category is coincidental or correlated, Concept Bottleneck Models (CBMs) gauge their impact on VLMs' decision-making, with high-influence ones being identified as spurious attributes. By leveraging \texttt{SAP}, a pure and robust attribute pool is achieved, significantly improving the generalization of existing attribute-based methods.

\textbf{Prompting MLLMs.} Here we assume $\mathcal{U}$ as a MLLM, differing from prior methods that only accept textual prompts. $\mathcal{U}$ concurrently processes both prompts and images as input such that $\mathcal{A}_{c} = \mathcal{U}(\mathcal{X}_{c}, \mathcal{H}(c))$, where $\mathcal{X}_{c}$ represents training images labeled with $c$, and $\mathcal{H}
(c)$ is a set of chain-of-thought prompts probing two aspects: core attributes and non-core counterparts. Specifically, we design three question formats:
\begin{center}
\begin{tabular}{l}
    \textbf{Q1:} List all the visual cues you see in the photo: \\
    \textbf{Q2:} Are the objects you list a part of \_\_\_? \\
    \textbf{Q3:} Describe \_\_\_ in the photo in details: \\
  \end{tabular}
\end{center} 
Combining Q1 and Q2 helps identify non-core attributes in the images, while Q3 provides detailed core attributes belonging to the category. Empirically, we'll use multiple templates for each question to ensure thoroughness. Upon reformulation, we derive $\mathcal{A}_{c} = \widetilde{\mathcal{A}}^{-}_{c}\cup \mathcal{A}^{+}_{c}$, where $\widetilde{\mathcal{A}}^{-}_{c}$ denotes non-core attributes, and $\mathcal{A}^{+}_{c}$ represents the core ones.

\textbf{Finding spurious attributes.} Given non-core attributes $\widetilde{\mathcal{A}}^{-}_{c}$, we use their weights on model predictions as a proxy to indicate the extent of spurious correlations. We use a CBM~\citep{koh2020concept} to achieve this goal. Specifically, we construct a bottleneck embedding $\mathcal{E} \in \mathbb{R}^{N \times d}$ against the attribute pool $\mathcal{P}$, where $N = \vert \mathcal{C} \vert \times J$ indicates the total number of attributes in the current pool, each row $\mathcal{E}_{i}\in \mathbb{R}^{d}$ indicates the feature of the corresponding attribute, and $d$ is the feature dimension. In other words, $\mathcal{E}$ is a feature matrix that concatenates attributes across all the categories together. The procedure of CBMs is to combine two functions to make the prediction: $\hat{c} = h(g(\phi_{I}(x), \mathcal{E}))$, where $g$: $\mathbb{R}^d\times\mathbb{R}^{N \times d}\to\mathbb{R}^N$ measures the score between the image feature and every attribute in the bottleneck, $h$: $\mathbb{R}^{N} \rightarrow \mathcal{C}$ produces the final prediction based on the score vector. Following \citet{yang2023language}, we set the score vector $g$ as the dot product of the features between two modalities $g(\phi_{I}(x), \mathcal{E}) = \phi_{I}(x) \cdot \mathcal{E}^{\intercal}$, and $h$ as the linear projection with a learnable weight matrix $\mathcal{W} \in \mathbb{R}^{\vert \mathcal{C} \vert \times N}$ such that $h(g;\mathcal{W}) = softmax(g \cdot \mathcal{W}^{\intercal})$. Intuitively, $\mathcal{W}_{ij}$ indicates the impact factor of $j^{th}$ attribute, \ie, $a_{i}^{j}$ on the prediction $i$. To learn the weights of attributes on predictions, a cross-entropy loss is typically employed. 

For non-core attributes, a high weight indicates a strong correlation with the category, whereas a low weight suggests that its presence might be coincidental. Thus, a natural idea to confirm spurious attributes is by thresholding $\gamma$. Formally, for a specific prediction $i$, the spurious attributes are defined as
\begin{equation}
\label{eq:spurious}
    \mathcal{A}_{i}^{-} = \{a_{i}^{j} \in \widetilde{\mathcal{A}}_{i}^{-} \ \vert \ \mathcal{W}_{ij} \geq \gamma\}.
\end{equation}

There is a trade-off in choosing $\gamma$. If $\gamma$ is too large, some spurious attributes may be overlooked. Conversely, if it is too small, a large number of false positives may be introduced. Additionally, we observe significant variability in attribute weight distributions among different categories, posing challenges in identifying spurious attributes with a uniform threshold. Creating a manual threshold for each category is prohibitively expensive. Hence, we introduce an adaptive strategy. Given a prediction $c$, we select $\gamma_{c}$ as the lowest weight of $\mathcal{A}_{c}^{+}$ such that non-core attributes with weights higher than any of the core attributes are considered spurious. This ensures flexible selection of spurious attributes, greatly aiding \texttt{SAS} introduced in Section~\hyperref[subsec:sas]{3.4} to be discussed next.
\subsection*{3.4 \quad Spurious Attribute Shielding}
\label{subsec:sas}
\begin{figure*}[t!] 
    \centering
    \includegraphics[width=1\linewidth]{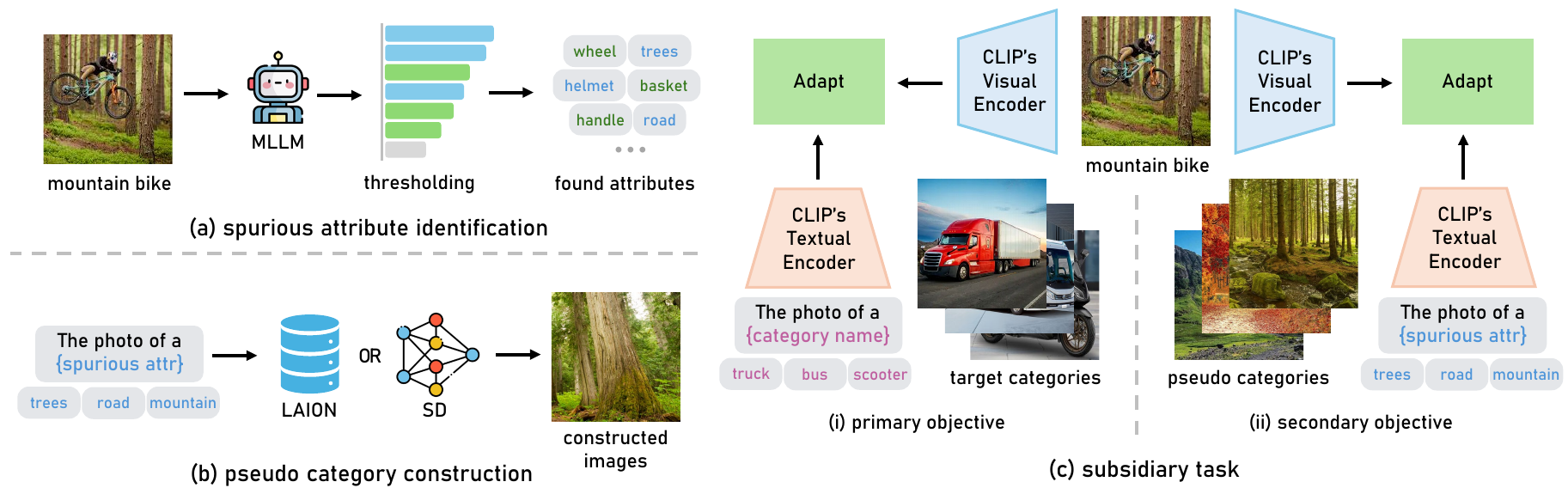}
    \caption{\textbf{The overview of \texttt{SAS}.} In \textbf{(a)}, we generate and identify spurious attributes with \texttt{SAP}. In \textbf{(b)}, we construct pseudo categories by synthetic data (SD) or retrieval (LAION). In \textbf{(c)}, apart from the main objective (i), \eg, cross-entropy loss, we introduce an auxiliary subsidiary task (ii) for learning robust features.
    }
\label{fig:SaS}
\vspace{-6mm}
\end{figure*}
\texttt{SAP} complements existing attribute-based methods by screening out spurious attributes from the pool, while it does not prevent the PEFT family from learning spurious features in the training images. Hence, we propose {\sc Spurious Attribute Shielding} (\texttt{SAS}), a plug-and-play module to be seamlessly integrated into arbitrary PEFT methods by mitigating the influence of spurious features. Building upon the spurious attributes detected by \texttt{SAP}, \texttt{SAS} introduces a subsidiary task by constructing a set of pseudo categories alongside the real one and let VLMs differentiate among them. This auxiliary learning objective effectively prompts VLMs to learn robust features rather than ones referred by these spurious attributes. For instance, if ${\rm streetlight}$ is a spurious attribute for the category ${\rm vehicle}$ impacting decision-making significantly, we introduce a pseudo category specifically for ${\rm streetlight}$ and differentiate between the two, thereby reducing the reliance of ${\rm streetlight}$ when identifying ${\rm vehicle}$. Fig.~\ref{fig:SaS} demonstrates the overall pipeline of \texttt{SAS}.

Formally, given a category $c$, we establish a set of pseudo categories $\mathcal{J}_{c}$ with constructed images $\{\hat{\mathcal{X}}_{j} \mid{j \in \mathcal{J}_{c}}\}$. Thus we define a subsidiary dataset $\mathcal{D}_{c} = \{(x, y)\mid x \in \hat{\mathcal{X}} \cup \mathcal{X}_{c} \textrm{ and } y \in \mathcal{J}_{c} \cup \{c\}\}$. We aim to optimize the following
\begin{equation}
\label{eq:pse}
     \mathcal{L}_{pse} = -\sum\limits_{c\in\mathcal{C}}\mathop{\mathbb{E}}\limits_{(x, y) \in \mathcal{D}_{c}}\log \frac{\exp(s(\phi_{I}(x), \phi_{L}(t_{y})) / \tau)}{\sum\limits_{j \in \mathcal{J}_{c} \cup \{c\}}\exp(s(\phi_{I}(x), \phi_{L}(t_{j})) / \tau)}.
\end{equation}
That is, we introduce an additional cross-entropy loss for classifying between each target category and its corresponding pseudo categories, which are defined by spurious attributes. This can be viewed as a subsidiary task, aimed at reducing reliance on spurious attributes while achieving correct classification in the downstream task. When integrated with existing methods, we introduce a scaler $\lambda$ to balance the importance of $\mathcal{L}_{pse}$: $\mathcal{L}_{tot} = \mathcal{L}_{ce} + \lambda \mathcal{L}_{pse}$. 

A natural question to ask is: how to construct $\hat{\mathcal{X}}$ such that the adapted modules may effectively distinguish spurious attributes from the target categories? In this work, we introduce two approaches.

\textbf{Synthetic Generation.} 
We create pseudo categories using synthetic data by leveraging the text-to-image model Stable Diffusion (SD)~\citep{rombach2022high}. We consider two key factors: 1) diversity: Our goal is for pseudo categories to fully represent the features of spurious attributes. To achieve this, we use LLMs to generate various prompts, which are then used as inputs to SD to produce a range of images. 2) purity: If the constructed images contain not only spurious attributes but also unexpected elements, \ie, noise attributes, these noise attributes may create new shortcuts, affecting the effectiveness of \texttt{SAS}. Empirically, selecting the top-k images that are most similar to the corresponding spurious attribute can help reduce the presence of noise attributes. Further details are in Supp. Mat. \hyperref[subsec:A4]{A}.

\textbf{Pretraining Retrieval.} 
An alternative way is to gather image samples from pre-training data such as LAION-5B~\citep{schuhmann2022laion}, a publicly available subset of CLIP's pre-training datasets. We use captions as a proxy to efficiently determine semantic similarity between pre-training images and spurious attributes. Finally, we select the top-k matches to the spurious attributes to create the pseudo categories. 
\section*{4 \quad Experiment}
\label{sec:experiment}
\begin{figure*}[t!] 
    \centering
    \includegraphics[width=1\linewidth]{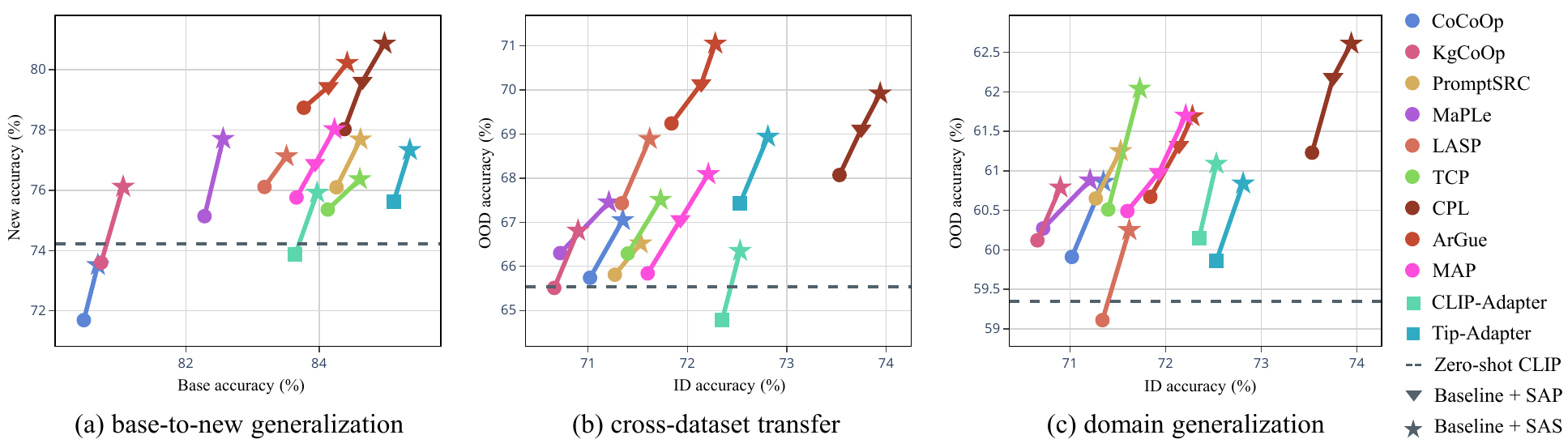}
    \caption{\textbf{The average results of three generalization tasks over 11 datasets.} The x-axis and y-axis represent in-distribution/base accuracy and out-of-distribution/new accuracy, respectively. We present the out-of-distribution accuracy of vanilla CLIP as a horizontal bar to represent the zero-shot capability. The detailed numerical results are provided in Supp. Mat. \hyperref[subsec:E4]{E}.
    }
\label{fig:main_results}
\vspace{-4mm}
\end{figure*}
\begin{figure}
\begin{minipage}{0.40\textwidth}
\centering
\includegraphics[width=1\linewidth]{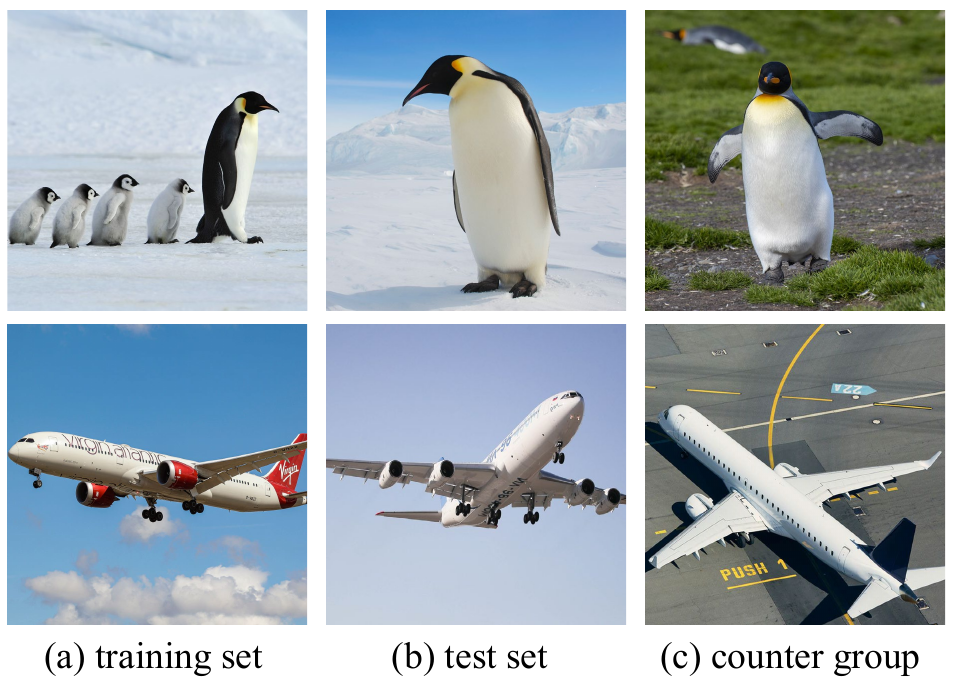}
\caption{\textbf{Example samples from test set and counter group.} The samples from counter group do not contain spurious attributes, \eg, ${\rm ice}$ or ${\rm sky}$.}
\label{fig: counter}
\end{minipage}
\hspace*{0.2cm}
\begin{minipage}{0.55\textwidth}
\captionsetup{type=table} 
\footnotesize
\centering
\setlength{\tabcolsep}{0.8mm}
\renewcommand{\arraystretch}{1.315}
\begin{tabular}{lcccccc}
\ChangeRT{1.2pt}
\multicolumn{1}{c}{\multirow{2}{*}{Method}} & \multicolumn{2}{c}{ImageNet}    & \multicolumn{2}{c}{OxfordPets}  & \multicolumn{2}{c}{FGVCAircraft} \\
\multicolumn{1}{c}{}                        & Test           & Counter        & Test           & Counter        & Test            & Counter        \\ \hline
PSRC                                        & 72.46          & 63.06          & 87.51          & 74.08          & \textbf{39.72}  & 21.35          \\
PSRC + \texttt{SAS}                                  & \textbf{73.82} & \textbf{68.55} & \textbf{88.27} & \textbf{77.17} & 39.54           & \textbf{26.43} \\ \hline
TCP                                         & 71.54          & 60.17          & 89.11          & 72.84          & 38.30           & 23.61          \\
TCP + \texttt{SAS}                                   & \textbf{72.81} & \textbf{64.50} & \textbf{89.86} & \textbf{76.14} & \textbf{39.23}  & \textbf{28.44} \\ \hline
CPL                                         & 69.08          & 64.75          & \textbf{90.32} & 79.91          & 40.43           & 27.65          \\
CPL + \texttt{SAS}                                   & \textbf{70.15} & \textbf{67.98} & 90.12          & \textbf{83.45} & \textbf{40.64}  & \textbf{32.12} \\ \ChangeRT{1.2pt}
\end{tabular}
\caption{\textbf{The results for standard few-shot classification on test set and counter group, respectively.} Essentially, counter group is a subset of test set where spurious attributes are removed.}
\label{tab:counter}
\end{minipage}
\vspace{-4mm}
\end{figure}
\textbf{Task Setting.} Following previous work~\citep{MaPLe, zhou2022conditional, khattak2023self}, the experiment is conducted on base-to-new generalization, cross-dataset transfer and domain generalization. 
For base-to-new generalization, the datasets are equally divided into base and new categories, where the model is trained on base categories and evaluated on unseen ones. For cross-dataset transfer, the model will be trained on a large-scale dataset, and generalized across various other datasets. For domain generalization, the model will be transferred from an in-distribution dataset to several variants.

\textbf{Datasets.} For base-to-new generalization, we employ 11 datasets, including ImageNet~\citep{deng2009imagenet}, Caltech101~\citep{fei2004learning}, OxfordPets~\citep{parkhi12a}, StanfordCars~\citep{krause20133d}, Flowers102~\citep{nilsback2008automated}, Food101~\citep{bossard2014food}, FGVCAircraft~\citep{maji2013fine}, SUN397~\citep{xiao2010sun}, UCF101~\citep{soomro2012dataset}, DTD~\citep{cimpoi2014describing} and EuroSAT~\citep{helber2019eurosat}. For cross-dataset transfer, we train models on ImageNet~\citep{deng2009imagenet}, and evaluate on the remaining datasets mentioned above. For domain generalization, we designate ImageNet as the in-distribution dataset, with four out-of-distribution variants encompassing ImageNetV2~\citep{recht2019imagenet}, ImageNet-Sketch~\citep{wang2019learning}, ImageNet-A~\citep{hendrycks2021natural} and ImageNet-R~\citep{hendrycks2021many}. The experiments 
are carried out in the few-shot setting, where we randomly sample 16 shots for each category to compose the training set. 

\textbf{Baselines.} We consider various PEFT approaches. Specifically, for prompt tuning, we consider category conditioning including CoCoOp~\citep{zhou2022conditional} and TCP~\citep{, yao2023tcp}, regularization techniques encompassing KgCoOp~\citep{yao2023visual}, LASP~\citep{bulat2023lasp} and PromptSRC~\citep{khattak2023self}, attribute-based methods such as CPL~\citep{zhang2024concept}, ArGue~\citep{tian2023argue} and MAP~\citep{liu2024multi}. We also consider multi-modal prompt tuning, \ie, MaPLe~\citep{MaPLe}. Besides, CLIP-Adapter~\citep{gao2024clip} and its training-free version Tip-Adapter~\citep{zhang2021tip} are involved. All results are averaged over three runs with distinct initialization.

\textbf{Implementation Details.} 
Unless specified otherwise, we use pre-trained CLIP~\citep{radford2021learning} and ViT-B16 as the visual backbone for fair comparison. Since our proposed method is a plug-and-play module, we strictly adhere to the settings of existing works, including optimizers, batch size, learning rate, and other strategies. This indicates that for different baselines, we may use distinct hyperparameters specified in their respective papers. For \texttt{SAP}, we use GPT-4V Turbo~\citep{achiam2023gpt} as the MLLM, with a temperature scaler of $0.7$ and an image understanding level set to \textit{high}. For \texttt{SAS}, by default, we use ChatGPT to generate 5 prompts for each spurious attribute, which are then fed into Stable Diffusion~\citep{rombach2022high} to create pseudo categories. More details, such as the effect of choices of MLLMs and comparison between pseudo category construction with synthesized and pre-training data, are provided in Supp. Mat. \hyperref[subsec:B4]{B}. Each pseudo category contains 16 shots, matching the number in the target category. All experiments are conducted on a single NVIDIA 4090 GPU.
\subsection*{4.1 \quad Main Results}
\label{subsec:main}
\begin{figure*}[t!] 
    \centering
    \includegraphics[width=1\linewidth]{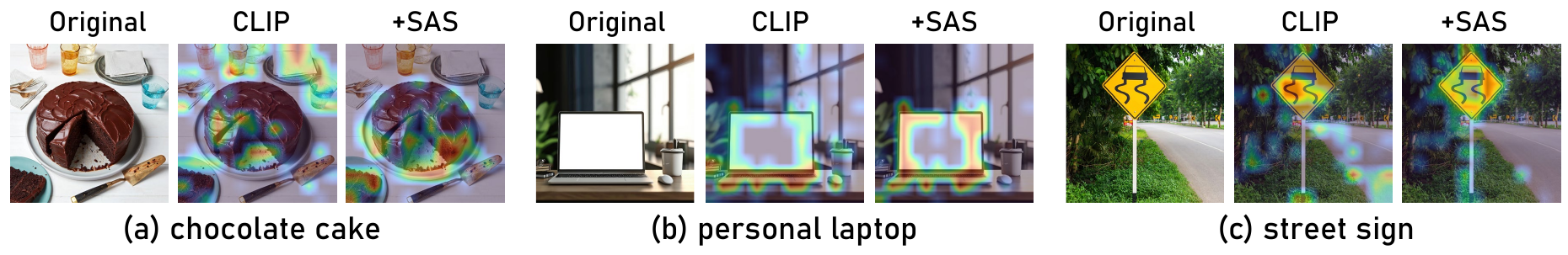}
    \caption{\textbf{The saliency map of VLMs with and without \texttt{SAS}.} From left to right we show three example categories, including ${\rm chocolate \ cake}$, ${\rm personal \ laptop}$, and ${\rm street \ sign}$.
    }
\label{fig:saliency}
\vspace{-0.5em}
\end{figure*}
\begin{figure}[!t]
\begin{minipage}[t]{\textwidth}
\begin{minipage}[t]{0.49\textwidth}
\makeatletter\def\@captype{table}
\footnotesize
\setlength{\tabcolsep}{0.8mm}
\renewcommand{\arraystretch}{1.1}
\begin{tabular}{cccccccc}
\ChangeRT{1.2pt}
\#p  & 1     & 2     & 3     & 4     & 5              & 6              & 7     \\ \hline
Base & 83.10 & 83.47 & 83.41 & 83.59 & 83.64 & 83.51          & \textbf{83.68} \\
New  & 75.05 & 75.89 & 76.91 & 77.19 & 77.36          & \textbf{77.41} & 77.23 \\
HM   & 78.87 & 79.50 & 80.03 & 80.26 & \textbf{80.38} & 80.34          & 80.32 \\ \ChangeRT{1.2pt}
\end{tabular}
\caption{\textbf{Varying number of SD prompts on base-to-new generalization.} The results are averaged across 11 datasets and 11 baselines.
} 
\label{tab:p}
\end{minipage}
\hspace*{0.05cm}
\begin{minipage}[t]{0.49\textwidth}
\makeatletter\def\@captype{table}
\footnotesize
\setlength{\tabcolsep}{0.8mm}
\renewcommand{\arraystretch}{1.1}
\begin{tabular}{cccccccc}
\ChangeRT{1.2pt}
$\gamma$ & 0.0   & 0.2   & 0.4   & 0.6   & 0.8   & 1.0   & Ada.           \\ \hline
Base  & 82.95 & 83.16 & 83.53 & 83.43 & 83.28 & 83.23 & \textbf{83.64} \\
New   & 76.21 & 76.60 & 76.41 & 76.01 & 75.30 & 74.82 & \textbf{77.36} \\
HM    & 79.44 & 79.75 & 79.81 & 79.55 & 79.09 & 78.80 & \textbf{80.38} \\ \ChangeRT{1.2pt}
\end{tabular}
\caption{\textbf{Varying $\gamma$ on base-to-new generalization.} We experiment with fixed values, juxtaposing them against the suggested adaptive strategy. 
}
\label{tab:gamma}
\end{minipage}
\vspace{-0.7em}
\end{minipage}
\end{figure}
\begin{table}[t!]
 \begin{minipage}[c]{0.37\textwidth}
    \footnotesize
    \setlength{\tabcolsep}{1.5mm}
    \renewcommand{\arraystretch}{1.1}
    \begin{tabular}{lcccc}
    \ChangeRT{1.2pt}
    \multicolumn{1}{c}{Method} & Epoch & Time $\downarrow$  & Accuracy $\uparrow$ & Gain $\uparrow$  \\ \hline
    ZSCLIP                     & \NA      &  \NA     & 70.22    &   \NA    \\ \hline
    CoCoOp                     & 10    & 4h37m & 73.10    &   \NA    \\
    \qquad + \texttt{SAS}                       & 10    & 6h18m & 74.21    & +1.11 \\
    \qquad + selective trick                  & 10    & 4h51m & 74.02    & +0.92 \\ \hline
    PromptSRC                       & 50    & 3h29m & 74.01    &    \NA   \\
    \qquad + \texttt{SAS}                       & 50    & 4h56m & 75.46    & +1.45 \\
    \qquad + selective trick                  & 50    & 3h38m & 75.20    & +1.19 \\ \ChangeRT{1.2pt}
    \end{tabular}
  \end{minipage}
  \hfill
\begin{minipage}[c]{0.39\textwidth}
    \caption{\textbf{The efficiency of \texttt{SAS} with and without selective trick.} We evaluate in terms of training time and accuracy gains given the same number of epochs. We opt for two time-intensive baselines, \ie, CoCoOp and PromptSRC, and train both on ImageNet under base-to-new generalization task.}
    \label{tab:trick}
    \vspace{-2mm}
  \end{minipage}
  \vspace{-4mm}
\end{table}
\textbf{Our method is complementary to PEFT approaches.}
Fig.~\ref{fig:main_results} depicts the results of baselines and their integration with \texttt{SAP} and \texttt{SAS}. We observe an upward trend in accuracy, indicating improvements in out-of-distribution accuracy without compromising downstream task performance. For conventional methods, \eg, CoCoOp, \texttt{SAS} helps achieve zero-shot capability on distribution shifts. For strong baselines, \eg, CPL, the incorporation of \texttt{SAP} and \texttt{SAS} enables them to reach a new state-of-the-art benchmark. Overall, applying our method leads to an average improvement of over 2\% in most baselines. These promising results align with the observation of the biased nature of VLMs, as demonstrated in Section~\hyperref[subsec:motivation]{3.2}.

\textbf{Our method is effective on counter test samples.}
To further highlight the effectiveness of \texttt{SAS}, we conduct standard few-shot classification in an adversarial evaluation manner. This involves selecting a subset from the original test set to create a counter group for evaluating the VLM. For each category, we filter out images from the test set that bear high semantic similarity to the identified spurious attributes, retaining only images free of such attributes. This counter group is significantly more challenging to predict using spurious attributes compared to the entire test set. Fig.~\ref{fig: counter} displays example images from the test set and counter group. Table~\ref{tab:counter} presents the improvement in accuracy of \texttt{SAS} over baselines, both for the test set and counter group. We notice that 1) for the counter group, its accuracy is much lower than that of the test set; 2) \texttt{SAS} effectively bridges this gap, with improvement on the counter group far exceeding that on the test set, up to approximately 6\%.
\subsection*{4.2 \quad Ablation Study}
\label{subsec:ablation}
\textbf{Diverse construction is beneficial for learning robust features.} During training, we aim for the constructed pseudo-categories to possess similar semantics to their target counterparts, thereby creating a strong contrast, while also maintaining high diversity to comprehensively represent spurious features. This trade-off is achieved by varying the number of SD prompts. As shown in Table~\ref{tab:p}, the effectiveness of \texttt{SAS} in assisting the baselines becomes more evident with an increasing number of prompts. This underscores the importance of the quality of pseudo-categories, which should thoroughly reflect the corresponding spurious attributes.

\textbf{Selecting appropriate spurious attributes matters.} The core principle of \texttt{SAS} is to introduce auxiliary categories to be trained alongside the main task, preventing the model from achieving high accuracy through spurious features. A natural concern is whether the model's gains are due to the introduction of additional data rather than an increase in robustness. In other words, does the model genuinely learn to distinguish spurious attributes from pseudo categories? We investigate this by adjusting the threshold $\gamma$ to control the presence of spurious attributes in pseudo categories. A higher $\gamma$ indicates a shortage of identified spurious attributes, while a lower $\gamma$ may introduce false positives. In Table~\ref{tab:gamma}, we observe that performance significantly drops when $\gamma$ is either too high or too low. This indicates that 1) spurious attributes play a crucial role in the contribution of \texttt{SAS}, and 2) the introduction of noisy attributes actually impairs the model's robustness. Additionally, the suggested adaptive strategy, which allows for flexible selection of spurious attributes, outperforms the pre-defined $\gamma$.
\subsection*{4.3 \quad Further Analysis}
\label{subsec:further}
\textbf{\texttt{SAS} corrects the preference of VLMs on spurious attributes.} 
To qualitatively assess \texttt{SAS}'s impact on VLMs, we present the saliency maps of VLMs with and without \texttt{SAS} in Fig.~\ref{fig:saliency}. Common spurious correlations can be observed, such as (a) ${\rm utensils}$ appearing alongside ${\rm chocolate \ cake}$, and (b) a ${\rm mouse}$ typically appearing with a ${\rm laptop}$. In critical applications, such as autonomous driving, (c) ${\rm road}$ tends to act as confounders for ${\rm street \ sign}$. \texttt{SAS} can effectively shift attention from these spurious attributes to the corresponding main objects. While we revisit Fig.~\ref{fig:intro}(c), this also aligns with the interpretation of CBMs that \texttt{SAS} suppresses the influence of spurious attributes on predictions.

\textbf{Balancing the trade-off between efficiency and effectiveness.} One potential concern with \texttt{SAS} is its impact on training efficiency. Applying a distinct loss to each category can be computationally demanding. To address this, we introduce a selective optimization trick. Rather than targeting all categories, we only optimize ones that heavily rely on spurious attributes for predictions. Details of this approach are outlined in Supp. Mat. \hyperref[subsec:C2]{C}. In Table~\ref{tab:trick}, we demonstrate the effectiveness of this selective strategy by optimizing only 10\% of the categories, showing the training time and accuracy. This approach significantly reduces \texttt{SAS}'s training time while preserving most of its accuracy gains.
\section*{5 \quad Conclusion}
\label{sec:conclusion}
This paper is motivated by an often-overlooked fact: VLMs tend to favor spurious attributes in their predictions, leading to decreased accuracy on out-of-distribution datasets. To tackle this issue, we first introduce {\sc Spurious Attribute Probing} (\texttt{SAP}), which identifies and filters out these problematic attributes, significantly improving the generalization of existing attribute-based methods. Furthermore, to alleviate the biased nature of VLMs, we introduce {\sc Spurious Attribute Shielding} (\texttt{SAS}), a plug-and-play module that reduces the influence of these attributes on predictions and complements various PEFT approaches. Both solutions significantly enhance accuracy in handling distribution shifts without compromising performance on downstream tasks, achieving a new state-of-the-art level.
\section*{6 \quad Acknowledgement}
This research was, in part, funded by the U.S. Government – DARPA TIAMAT HR00112490421. The views and conclusions contained in this document are those of the authors and should not be interpreted as representing the official policies, either expressed or implied, of the U.S. Government.
\bibliography{iclr2025_conference}

\begin{thebibliography}{92}
\providecommand{\natexlab}[1]{#1}
\providecommand{\url}[1]{\texttt{#1}}
\expandafter\ifx\csname urlstyle\endcsname\relax
  \providecommand{\doi}[1]{doi: #1}\else
  \providecommand{\doi}{doi: \begingroup \urlstyle{rm}\Url}\fi

\bibitem[Achiam et~al.(2023)Achiam, Adler, Agarwal, Ahmad, Akkaya, Aleman, Almeida, Altenschmidt, Altman, Anadkat, et~al.]{achiam2023gpt}
Josh Achiam, Steven Adler, Sandhini Agarwal, Lama Ahmad, Ilge Akkaya, Florencia~Leoni Aleman, Diogo Almeida, Janko Altenschmidt, Sam Altman, Shyamal Anadkat, et~al.
\newblock Gpt-4 technical report.
\newblock \emph{arXiv preprint arXiv:2303.08774}, 2023.

\bibitem[Adila et~al.(2023)Adila, Shin, Cai, and Sala]{adila2023zero}
Dyah Adila, Changho Shin, Linrong Cai, and Frederic Sala.
\newblock Zero-shot robustification of zero-shot models with foundation models.
\newblock In \emph{ICLR}, 2023.

\bibitem[An et~al.(2023)An, Zhu, Panaitescu-Liess, Mummadi, and Huang]{an2023more}
Bang An, Sicheng Zhu, Michael-Andrei Panaitescu-Liess, Chaithanya~Kumar Mummadi, and Furong Huang.
\newblock More context, less distraction: Improving zero-shot inference of clip by inferring and describing spurious features.
\newblock In \emph{Workshop on Efficient Systems for Foundation Models@ ICML2023}, 2023.

\bibitem[Berg et~al.(2022)Berg, Hall, Bhalgat, Kirk, Shtedritski, and Bain]{berg2022prompt}
Hugo Berg, Siobhan Hall, Yash Bhalgat, Hannah Kirk, Aleksandar Shtedritski, and Max Bain.
\newblock A prompt array keeps the bias away: Debiasing vision-language models with adversarial learning.
\newblock In Yulan He, Heng Ji, Sujian Li, Yang Liu, and Chua-Hui Chang (eds.), \emph{ACL}, 2022.

\bibitem[Bossard et~al.(2014)Bossard, Guillaumin, and Gool]{bossard2014food}
Lukas Bossard, Matthieu Guillaumin, and Luc~Van Gool.
\newblock Food-101 - mining discriminative components with random forests.
\newblock In \emph{{ECCV} {(6)}}, volume 8694 of \emph{Lecture Notes in Computer Science}, pp.\  446--461. Springer, 2014.

\bibitem[Bulat \& Tzimiropoulos(2023)Bulat and Tzimiropoulos]{bulat2023lasp}
Adrian Bulat and Georgios Tzimiropoulos.
\newblock {LASP:} text-to-text optimization for language-aware soft prompting of vision {\&} language models.
\newblock In \emph{{CVPR}}, pp.\  23232--23241. {IEEE}, 2023.

\bibitem[Chen et~al.(2024)Chen, Wu, Wang, Su, Chen, Xing, Zhong, Zhang, Zhu, Lu, et~al.]{chen2024internvl}
Zhe Chen, Jiannan Wu, Wenhai Wang, Weijie Su, Guo Chen, Sen Xing, Muyan Zhong, Qinglong Zhang, Xizhou Zhu, Lewei Lu, et~al.
\newblock Internvl: Scaling up vision foundation models and aligning for generic visual-linguistic tasks.
\newblock In \emph{CVPR}, pp.\  24185--24198, 2024.

\bibitem[Chuang et~al.(2023)Chuang, Jampani, Li, Torralba, and Jegelka]{chuang2023debiasing}
Ching-Yao Chuang, Varun Jampani, Yuanzhen Li, Antonio Torralba, and Stefanie Jegelka.
\newblock Debiasing vision-language models via biased prompts.
\newblock \emph{arXiv preprint arXiv:2302.00070}, 2023.

\bibitem[Cimpoi et~al.(2014)Cimpoi, Maji, Kokkinos, Mohamed, and Vedaldi]{cimpoi2014describing}
Mircea Cimpoi, Subhransu Maji, Iasonas Kokkinos, Sammy Mohamed, and Andrea Vedaldi.
\newblock Describing textures in the wild.
\newblock In \emph{{CVPR}}, pp.\  3606--3613. {IEEE} Computer Society, 2014.

\bibitem[Deng et~al.(2009)Deng, Dong, Socher, Li, Li, and Fei{-}Fei]{deng2009imagenet}
Jia Deng, Wei Dong, Richard Socher, Li{-}Jia Li, Kai Li, and Li~Fei{-}Fei.
\newblock Imagenet: {A} large-scale hierarchical image database.
\newblock In \emph{{CVPR}}, pp.\  248--255. {IEEE} Computer Society, 2009.

\bibitem[Dettmers et~al.(2024)Dettmers, Pagnoni, Holtzman, and Zettlemoyer]{dettmers2024qlora}
Tim Dettmers, Artidoro Pagnoni, Ari Holtzman, and Luke Zettlemoyer.
\newblock Qlora: Efficient finetuning of quantized llms.
\newblock In \emph{NeurIPS}, volume~36, 2024.

\bibitem[Fei{-}Fei et~al.(2004)Fei{-}Fei, Fergus, and Perona]{fei2004learning}
Li~Fei{-}Fei, Rob Fergus, and Pietro Perona.
\newblock Learning generative visual models from few training examples: An incremental bayesian approach tested on 101 object categories.
\newblock In \emph{{CVPR} Workshops}, pp.\  178. {IEEE} Computer Society, 2004.

\bibitem[Gao et~al.(2024)Gao, Geng, Zhang, Ma, Fang, Zhang, Li, and Qiao]{gao2024clip}
Peng Gao, Shijie Geng, Renrui Zhang, Teli Ma, Rongyao Fang, Yongfeng Zhang, Hongsheng Li, and Yu~Qiao.
\newblock Clip-adapter: Better vision-language models with feature adapters.
\newblock In \emph{IJCV}, 2024.

\bibitem[Goyal et~al.(2017)Goyal, Ebrahimi~Kahou, Michalski, Materzynska, Westphal, Kim, Haenel, Fruend, Yianilos, Mueller-Freitag, et~al.]{goyal2017something}
Raghav Goyal, Samira Ebrahimi~Kahou, Vincent Michalski, Joanna Materzynska, Susanne Westphal, Heuna Kim, Valentin Haenel, Ingo Fruend, Peter Yianilos, Moritz Mueller-Freitag, et~al.
\newblock The" something something" video database for learning and evaluating visual common sense.
\newblock In \emph{ICCV}, pp.\  5842--5850, 2017.

\bibitem[Han et~al.(2022)Han, Liang, Yang, Liu, Li, Bian, Zhao, Wu, Zhang, and Yao]{han2022umix}
Zongbo Han, Zhipeng Liang, Fan Yang, Liu Liu, Lanqing Li, Yatao Bian, Peilin Zhao, Bingzhe Wu, Changqing Zhang, and Jianhua Yao.
\newblock Umix: Improving importance weighting for subpopulation shift via uncertainty-aware mixup.
\newblock In \emph{NeurIPS}, volume~35, pp.\  37704--37718, 2022.

\bibitem[Helber et~al.(2019)Helber, Bischke, Dengel, and Borth]{helber2019eurosat}
Patrick Helber, Benjamin Bischke, Andreas Dengel, and Damian Borth.
\newblock Eurosat: {A} novel dataset and deep learning benchmark for land use and land cover classification.
\newblock \emph{{IEEE} J. Sel. Top. Appl. Earth Obs. Remote. Sens.}, 12\penalty0 (7):\penalty0 2217--2226, 2019.

\bibitem[Hendrycks et~al.(2021{\natexlab{a}})Hendrycks, Basart, Mu, Kadavath, Wang, Dorundo, Desai, Zhu, Parajuli, Guo, Song, Steinhardt, and Gilmer]{hendrycks2021many}
Dan Hendrycks, Steven Basart, Norman Mu, Saurav Kadavath, Frank Wang, Evan Dorundo, Rahul Desai, Tyler Zhu, Samyak Parajuli, Mike Guo, Dawn Song, Jacob Steinhardt, and Justin Gilmer.
\newblock The many faces of robustness: {A} critical analysis of out-of-distribution generalization.
\newblock In \emph{{ICCV}}, pp.\  8320--8329. {IEEE}, 2021{\natexlab{a}}.

\bibitem[Hendrycks et~al.(2021{\natexlab{b}})Hendrycks, Zhao, Basart, Steinhardt, and Song]{hendrycks2021natural}
Dan Hendrycks, Kevin Zhao, Steven Basart, Jacob Steinhardt, and Dawn Song.
\newblock Natural adversarial examples.
\newblock In \emph{{CVPR}}, pp.\  15262--15271. Computer Vision Foundation / {IEEE}, 2021{\natexlab{b}}.

\bibitem[Houlsby et~al.(2019)Houlsby, Giurgiu, Jastrzebski, Morrone, De~Laroussilhe, Gesmundo, Attariyan, and Gelly]{houlsby2019parameter}
Neil Houlsby, Andrei Giurgiu, Stanislaw Jastrzebski, Bruna Morrone, Quentin De~Laroussilhe, Andrea Gesmundo, Mona Attariyan, and Sylvain Gelly.
\newblock Parameter-efficient transfer learning for nlp.
\newblock In \emph{ICML}, 2019.

\bibitem[Hu et~al.(2021)Hu, Shen, Wallis, Allen-Zhu, Li, Wang, Wang, and Chen]{hu2021lora}
Edward~J Hu, Yelong Shen, Phillip Wallis, Zeyuan Allen-Zhu, Yuanzhi Li, Shean Wang, Lu~Wang, and Weizhu Chen.
\newblock Lora: Low-rank adaptation of large language models.
\newblock In \emph{ICLR}, 2021.

\bibitem[Huang et~al.(2024)Huang, Zhang, Jiang, and Lu]{huang2024open}
Jiaxing Huang, Jingyi Zhang, Kai Jiang, and Shijian Lu.
\newblock Open-vocabulary object detection via language hierarchy.
\newblock In \emph{NeurIPS}, 2024.

\bibitem[Kay et~al.(2017)Kay, Carreira, Simonyan, Zhang, Hillier, Vijayanarasimhan, Viola, Green, Back, Natsev, et~al.]{kay2017kinetics}
Will Kay, Joao Carreira, Karen Simonyan, Brian Zhang, Chloe Hillier, Sudheendra Vijayanarasimhan, Fabio Viola, Tim Green, Trevor Back, Paul Natsev, et~al.
\newblock The kinetics human action video dataset.
\newblock \emph{arXiv preprint arXiv:1705.06950}, 2017.

\bibitem[Khattak et~al.(2023{\natexlab{a}})Khattak, Rasheed, Maaz, Khan, and Khan]{MaPLe}
Muhammad~Uzair Khattak, Hanoona~Abdul Rasheed, Muhammad Maaz, Salman~H. Khan, and Fahad~Shahbaz Khan.
\newblock Maple: Multi-modal prompt learning.
\newblock In \emph{{CVPR}}, pp.\  19113--19122. {IEEE}, 2023{\natexlab{a}}.

\bibitem[Khattak et~al.(2023{\natexlab{b}})Khattak, Wasim, Naseer, Khan, Yang, and Khan]{khattak2023self}
Muhammad~Uzair Khattak, Syed~Talal Wasim, Muzammal Naseer, Salman Khan, Ming-Hsuan Yang, and Fahad~Shahbaz Khan.
\newblock Self-regulating prompts: Foundational model adaptation without forgetting.
\newblock In \emph{ICCV}, pp.\  15190--15200, 2023{\natexlab{b}}.

\bibitem[Kirillov et~al.(2023)Kirillov, Mintun, Ravi, Mao, Rolland, Gustafson, Xiao, Whitehead, Berg, Lo, et~al.]{kirillov2023segment}
Alexander Kirillov, Eric Mintun, Nikhila Ravi, Hanzi Mao, Chloe Rolland, Laura Gustafson, Tete Xiao, Spencer Whitehead, Alexander~C Berg, Wan-Yen Lo, et~al.
\newblock Segment anything.
\newblock In \emph{CVPR}, pp.\  4015--4026, 2023.

\bibitem[Koh et~al.(2020)Koh, Nguyen, Tang, Mussmann, Pierson, Kim, and Liang]{koh2020concept}
Pang~Wei Koh, Thao Nguyen, Yew~Siang Tang, Stephen Mussmann, Emma Pierson, Been Kim, and Percy Liang.
\newblock Concept bottleneck models.
\newblock In \emph{{ICML}}, volume 119 of \emph{Proceedings of Machine Learning Research}, pp.\  5338--5348. {PMLR}, 2020.

\bibitem[Krause et~al.(2013)Krause, Stark, Deng, and Fei{-}Fei]{krause20133d}
Jonathan Krause, Michael Stark, Jia Deng, and Li~Fei{-}Fei.
\newblock 3d object representations for fine-grained categorization.
\newblock In \emph{{ICCV} Workshops}, pp.\  554--561. {IEEE} Computer Society, 2013.

\bibitem[Krizhevsky et~al.(2009)Krizhevsky, Hinton, et~al.]{krizhevsky2009learning}
Alex Krizhevsky, Geoffrey Hinton, et~al.
\newblock Learning multiple layers of features from tiny images.
\newblock \emph{NA}, 2009.

\bibitem[Kuehne et~al.(2011)Kuehne, Jhuang, Garrote, Poggio, and Serre]{kuehne2011hmdb}
Hildegard Kuehne, Hueihan Jhuang, Est{\'\i}baliz Garrote, Tomaso Poggio, and Thomas Serre.
\newblock Hmdb: a large video database for human motion recognition.
\newblock In \emph{ICCV}, pp.\  2556--2563. IEEE, 2011.

\bibitem[Lester et~al.(2021)Lester, Al{-}Rfou, and Constant]{lester2021power}
Brian Lester, Rami Al{-}Rfou, and Noah Constant.
\newblock The power of scale for parameter-efficient prompt tuning.
\newblock In \emph{{EMNLP} {(1)}}, pp.\  3045--3059. Association for Computational Linguistics, 2021.

\bibitem[Li et~al.(2022{\natexlab{a}})Li, Li, Xiong, and Hoi]{li2022blip}
Junnan Li, Dongxu Li, Caiming Xiong, and Steven Hoi.
\newblock Blip: Bootstrapping language-image pre-training for unified vision-language understanding and generation.
\newblock In \emph{ICML}, 2022{\natexlab{a}}.

\bibitem[Li et~al.(2023{\natexlab{a}})Li, Li, Savarese, and Hoi]{li2023blip}
Junnan Li, Dongxu Li, Silvio Savarese, and Steven Hoi.
\newblock Blip-2: Bootstrapping language-image pre-training with frozen image encoders and large language models.
\newblock In \emph{ICML}, pp.\  19730--19742. PMLR, 2023{\natexlab{a}}.

\bibitem[Li et~al.(2022{\natexlab{b}})Li, Zhang, Zhang, Yang, Li, Zhong, Wang, Yuan, Zhang, Hwang, et~al.]{li2022grounded}
Liunian~Harold Li, Pengchuan Zhang, Haotian Zhang, Jianwei Yang, Chunyuan Li, Yiwu Zhong, Lijuan Wang, Lu~Yuan, Lei Zhang, Jenq-Neng Hwang, et~al.
\newblock Grounded language-image pre-training.
\newblock In \emph{CVPR}, pp.\  10965--10975, 2022{\natexlab{b}}.

\bibitem[Li et~al.(2024)Li, Wang, and Xie]{li2024inverse}
Xianhang Li, Zeyu Wang, and Cihang Xie.
\newblock An inverse scaling law for clip training.
\newblock In \emph{NeurIPS}, volume~36, 2024.

\bibitem[Li et~al.(2023{\natexlab{b}})Li, Fan, Hu, Feichtenhofer, and He]{li2023scaling}
Yanghao Li, Haoqi Fan, Ronghang Hu, Christoph Feichtenhofer, and Kaiming He.
\newblock Scaling language-image pre-training via masking.
\newblock In \emph{CVPR}, pp.\  23390--23400, 2023{\natexlab{b}}.

\bibitem[Li et~al.(2022{\natexlab{c}})Li, Hoogs, and Xu]{li2022discover}
Zhiheng Li, Anthony Hoogs, and Chenliang Xu.
\newblock Discover and mitigate unknown biases with debiasing alternate networks.
\newblock In \emph{ECCV}, pp.\  270--288. Springer, 2022{\natexlab{c}}.

\bibitem[Liao et~al.(2024)Liao, Tsiligkaridis, and Kulis]{liao2023descriptor}
Christopher Liao, Theodoros Tsiligkaridis, and Brian Kulis.
\newblock Descriptor and word soups: Overcoming the parameter efficiency accuracy tradeoff for out-of-distribution few-shot learning.
\newblock In \emph{CVPR}, 2024.

\bibitem[Liu et~al.(2024{\natexlab{a}})Liu, Li, Wu, and Lee]{liu2024visual}
Haotian Liu, Chunyuan Li, Qingyang Wu, and Yong~Jae Lee.
\newblock Visual instruction tuning.
\newblock In \emph{NeurIPS}, volume~36, 2024{\natexlab{a}}.

\bibitem[Liu et~al.(2021)Liu, Ji, Fu, Tam, Du, Yang, and Tang]{liu2021p}
Xiao Liu, Kaixuan Ji, Yicheng Fu, Weng~Lam Tam, Zhengxiao Du, Zhilin Yang, and Jie Tang.
\newblock P-tuning v2: Prompt tuning can be comparable to fine-tuning universally across scales and tasks.
\newblock In \emph{ACL}, 2021.

\bibitem[Liu et~al.(2024{\natexlab{b}})Liu, Wu, and Zhang]{liu2024multi}
Xin Liu, Jiamin Wu, and Tianzhu Zhang.
\newblock Multi-modal attribute prompting for vision-language models.
\newblock \emph{arXiv preprint arXiv:2403.00219}, 2024{\natexlab{b}}.

\bibitem[Liu et~al.(2018)Liu, Luo, Wang, and Tang]{liu2018large}
Ziwei Liu, Ping Luo, Xiaogang Wang, and Xiaoou Tang.
\newblock Large-scale celebfaces attributes (celeba) dataset.
\newblock \emph{Retrieved August}, 15\penalty0 (2018):\penalty0 11, 2018.

\bibitem[Long et~al.(2022)Long, Yin, Ajanthan, Nguyen, Purkait, Garg, Blair, Shen, and van~den Hengel]{long2022retrieval}
Alexander Long, Wei Yin, Thalaiyasingam Ajanthan, Vu~Nguyen, Pulak Purkait, Ravi Garg, Alan Blair, Chunhua Shen, and Anton van~den Hengel.
\newblock Retrieval augmented classification for long-tail visual recognition.
\newblock In \emph{CVPR}, pp.\  6959--6969, 2022.

\bibitem[Lugmayr et~al.(2022)Lugmayr, Danelljan, Romero, Yu, Timofte, and Van~Gool]{lugmayrinpainting}
Andreas Lugmayr, Martin Danelljan, Andres Romero, Fisher Yu, Radu Timofte, and L~Repaint Van~Gool.
\newblock Inpainting using denoising diffusion probabilistic models.
\newblock In \emph{CVPR}, pp.\  11461--11471, 2022.

\bibitem[Ma et~al.(2023)Ma, Yang, Ju, Zhang, Zhang, and Wang]{ma2023attrseg}
Chaofan Ma, Yuhuan Yang, Chen Ju, Fei Zhang, Ya~Zhang, and Yanfeng Wang.
\newblock Attrseg: Open-vocabulary semantic segmentation via attribute decomposition-aggregation.
\newblock In \emph{NeurIPS}, 2023.

\bibitem[Maji et~al.(2013)Maji, Rahtu, Kannala, Blaschko, and Vedaldi]{maji2013fine}
Subhransu Maji, Esa Rahtu, Juho Kannala, Matthew~B. Blaschko, and Andrea Vedaldi.
\newblock Fine-grained visual classification of aircraft.
\newblock \emph{CoRR}, abs/1306.5151, 2013.

\bibitem[Menon \& Vondrick(2023)Menon and Vondrick]{menon2022visual}
Sachit Menon and Carl Vondrick.
\newblock Visual classification via description from large language models.
\newblock In \emph{{ICLR}}. OpenReview.net, 2023.

\bibitem[Nilsback \& Zisserman(2008)Nilsback and Zisserman]{nilsback2008automated}
Maria{-}Elena Nilsback and Andrew Zisserman.
\newblock Automated flower classification over a large number of classes.
\newblock In \emph{{ICVGIP}}, pp.\  722--729. {IEEE} Computer Society, 2008.

\bibitem[Parashar et~al.(2024)Parashar, Lin, Liu, Dong, Li, Ramanan, Caverlee, and Kong]{parashar2024neglected}
Shubham Parashar, Zhiqiu Lin, Tian Liu, Xiangjue Dong, Yanan Li, Deva Ramanan, James Caverlee, and Shu Kong.
\newblock The neglected tails of vision-language models.
\newblock In \emph{CVPR}, 2024.

\bibitem[Parkhi et~al.(2012)Parkhi, Vedaldi, Zisserman, and Jawahar]{parkhi12a}
Omkar~M. Parkhi, Andrea Vedaldi, Andrew Zisserman, and C.~V. Jawahar.
\newblock Cats and dogs.
\newblock In \emph{{CVPR}}, pp.\  3498--3505. {IEEE} Computer Society, 2012.

\bibitem[Pratt et~al.(2023)Pratt, Covert, Liu, and Farhadi]{pratt2023does}
Sarah Pratt, Ian Covert, Rosanne Liu, and Ali Farhadi.
\newblock What does a platypus look like? generating customized prompts for zero-shot image classification.
\newblock In \emph{ICCV}, pp.\  15691--15701, 2023.

\bibitem[Radford et~al.(2021)Radford, Kim, Hallacy, Ramesh, Goh, Agarwal, Sastry, Askell, Mishkin, Clark, Krueger, and Sutskever]{radford2021learning}
Alec Radford, Jong~Wook Kim, Chris Hallacy, Aditya Ramesh, Gabriel Goh, Sandhini Agarwal, Girish Sastry, Amanda Askell, Pamela Mishkin, Jack Clark, Gretchen Krueger, and Ilya Sutskever.
\newblock Learning transferable visual models from natural language supervision.
\newblock In \emph{{ICML}}, volume 139 of \emph{Proceedings of Machine Learning Research}, pp.\  8748--8763. {PMLR}, 2021.

\bibitem[Rasheed et~al.(2023)Rasheed, Khattak, Maaz, Khan, and Khan]{rasheed2023fine}
Hanoona Rasheed, Muhammad~Uzair Khattak, Muhammad Maaz, Salman Khan, and Fahad~Shahbaz Khan.
\newblock Fine-tuned clip models are efficient video learners.
\newblock In \emph{CVPR}, pp.\  6545--6554, 2023.

\bibitem[Recht et~al.(2019)Recht, Roelofs, Schmidt, and Shankar]{recht2019imagenet}
Benjamin Recht, Rebecca Roelofs, Ludwig Schmidt, and Vaishaal Shankar.
\newblock Do imagenet classifiers generalize to imagenet?
\newblock In \emph{{ICML}}, volume~97 of \emph{Proceedings of Machine Learning Research}, pp.\  5389--5400. {PMLR}, 2019.

\bibitem[Rombach et~al.(2022)Rombach, Blattmann, Lorenz, Esser, and Ommer]{rombach2022high}
Robin Rombach, Andreas Blattmann, Dominik Lorenz, Patrick Esser, and Bj{\"o}rn Ommer.
\newblock High-resolution image synthesis with latent diffusion models.
\newblock In \emph{CVPR}, 2022.

\bibitem[Roth et~al.(2023)Roth, Kim, Koepke, Vinyals, Schmid, and Akata]{roth2023waffling}
Karsten Roth, Jae~Myung Kim, A.~Sophia Koepke, Oriol Vinyals, Cordelia Schmid, and Zeynep Akata.
\newblock Waffling around for performance: Visual classification with random words and broad concepts.
\newblock In \emph{ICCV}, pp.\  15746--15757, October 2023.

\bibitem[Sagawa et~al.(2019)Sagawa, Koh, Hashimoto, and Liang]{sagawa2019distributionally}
Shiori Sagawa, Pang~Wei Koh, Tatsunori~B Hashimoto, and Percy Liang.
\newblock Distributionally robust neural networks for group shifts: On the importance of regularization for worst-case generalization.
\newblock \emph{arXiv preprint arXiv:1911.08731}, 2019.

\bibitem[Santurkar et~al.(2020)Santurkar, Tsipras, and Madry]{santurkar2020breeds}
Shibani Santurkar, Dimitris Tsipras, and Aleksander Madry.
\newblock Breeds: Benchmarks for subpopulation shift.
\newblock In \emph{ICLR}, 2020.

\bibitem[Schuhmann et~al.(2022)Schuhmann, Beaumont, Vencu, Gordon, Wightman, Cherti, Coombes, Katta, Mullis, Wortsman, et~al.]{schuhmann2022laion}
Christoph Schuhmann, Romain Beaumont, Richard Vencu, Cade Gordon, Ross Wightman, Mehdi Cherti, Theo Coombes, Aarush Katta, Clayton Mullis, Mitchell Wortsman, et~al.
\newblock Laion-5b: An open large-scale dataset for training next generation image-text models.
\newblock In \emph{NeurIPS}, 2022.

\bibitem[Seth et~al.(2023)Seth, Hemani, and Agarwal]{seth2023dear}
Ashish Seth, Mayur Hemani, and Chirag Agarwal.
\newblock Dear: Debiasing vision-language models with additive residuals.
\newblock In \emph{CVPR}, 2023.

\bibitem[Silva-Rodriguez et~al.(2024)Silva-Rodriguez, Hajimiri, Ben~Ayed, and Dolz]{silva2024closer}
Julio Silva-Rodriguez, Sina Hajimiri, Ismail Ben~Ayed, and Jose Dolz.
\newblock A closer look at the few-shot adaptation of large vision-language models.
\newblock In \emph{CVPR}, pp.\  23681--23690, 2024.

\bibitem[Singla \& Feizi(2021)Singla and Feizi]{singla2021salient}
Sahil Singla and Soheil Feizi.
\newblock Salient imagenet: How to discover spurious features in deep learning?
\newblock In \emph{ICLR}, 2021.

\bibitem[Soomro(2012)]{soomro2012ucf101}
K~Soomro.
\newblock Ucf101: A dataset of 101 human actions classes from videos in the wild.
\newblock \emph{arXiv preprint arXiv:1212.0402}, 2012.

\bibitem[Soomro et~al.(2012)Soomro, Zamir, and Shah]{soomro2012dataset}
Khurram Soomro, Amir~Roshan Zamir, and Mubarak Shah.
\newblock A dataset of 101 human action classes from videos in the wild.
\newblock \emph{Center for Research in Computer Vision}, 2\penalty0 (11), 2012.

\bibitem[Sun et~al.(2023)Sun, Fang, Wu, Wang, and Cao]{sun2023eva}
Quan Sun, Yuxin Fang, Ledell Wu, Xinlong Wang, and Yue Cao.
\newblock Eva-clip: Improved training techniques for clip at scale.
\newblock \emph{arXiv preprint arXiv:2303.15389}, 2023.

\bibitem[Sung et~al.(2022)Sung, Cho, and Bansal]{sung2022vl}
Yi-Lin Sung, Jaemin Cho, and Mohit Bansal.
\newblock Vl-adapter: Parameter-efficient transfer learning for vision-and-language tasks.
\newblock In \emph{CVPR}, pp.\  5227--5237, 2022.

\bibitem[Teotia et~al.(2022)Teotia, Mao, and Vondrick]{teotia2022finding}
Revant Teotia, Chengzhi Mao, and Carl Vondrick.
\newblock Finding spuriously correlated visual attributes.
\newblock In \emph{ICML 2022: Workshop on Spurious Correlations, Invariance and Stability}, 2022.

\bibitem[Tian et~al.(2024)Tian, Zou, Yang, and Zhang]{tian2023argue}
Xinyu Tian, Shu Zou, Zhaoyuan Yang, and Jing Zhang.
\newblock Argue: Attribute-guided prompt tuning for vision-language models.
\newblock In \emph{CVPR}, 2024.

\bibitem[Udandarao et~al.(2022)Udandarao, Gupta, and Albanie]{udandarao2022sus}
Vishaal Udandarao, Ankush Gupta, and Samuel Albanie.
\newblock Sus-x: Training-free name-only transfer of vision-language models.
\newblock In \emph{ICCV}, 2022.

\bibitem[Utama et~al.(2020)Utama, Moosavi, and Gurevych]{utama2020towards}
Prasetya~Ajie Utama, Nafise~Sadat Moosavi, and Iryna Gurevych.
\newblock Towards debiasing nlu models from unknown biases.
\newblock In \emph{EMNLP}, 2020.

\bibitem[Wang et~al.(2015)Wang, Shen, Shao, Zhang, Xue, and Zhang]{wang2015multiple}
Dequan Wang, Zhiqiang Shen, Jie Shao, Wei Zhang, Xiangyang Xue, and Zheng Zhang.
\newblock Multiple granularity descriptors for fine-grained categorization.
\newblock In \emph{ICCV}, pp.\  2399--2406, 2015.

\bibitem[Wang et~al.(2019)Wang, Ge, Lipton, and Xing]{wang2019learning}
Haohan Wang, Songwei Ge, Zachary~C. Lipton, and Eric~P. Xing.
\newblock Learning robust global representations by penalizing local predictive power.
\newblock In \emph{NeurIPS}, pp.\  10506--10518, 2019.

\bibitem[Wang et~al.(2024)Wang, Lin, Chen, Schmidt, Han, and Zhang]{wang2024clips}
Qizhou Wang, Yong Lin, Yongqiang Chen, Ludwig Schmidt, Bo~Han, and Tong Zhang.
\newblock Do clips always generalize better than imagenet models?
\newblock \emph{arXiv preprint arXiv:2403.11497}, 2024.

\bibitem[Wei et~al.(2019)Wei, Yang, Wang, Deng, and Liu]{wei2019adversarial}
Kun Wei, Muli Yang, Hao Wang, Cheng Deng, and Xianglong Liu.
\newblock Adversarial fine-grained composition learning for unseen attribute-object recognition.
\newblock In \emph{ICCV}, pp.\  3741--3749, 2019.

\bibitem[Wong et~al.(2021)Wong, Santurkar, and Madry]{wong2021leveraging}
Eric Wong, Shibani Santurkar, and Aleksander Madry.
\newblock Leveraging sparse linear layers for debuggable deep networks.
\newblock In \emph{ICML}, pp.\  11205--11216, 2021.

\bibitem[Wu et~al.(2023)Wu, Yuksekgonul, Zhang, and Zou]{wu2023discover}
Shirley Wu, Mert Yuksekgonul, Linjun Zhang, and James Zou.
\newblock Discover and cure: Concept-aware mitigation of spurious correlation.
\newblock In \emph{ICML}, pp.\  37765--37786. PMLR, 2023.

\bibitem[Xiao et~al.(2010)Xiao, Hays, Ehinger, Oliva, and Torralba]{xiao2010sun}
Jianxiong Xiao, James Hays, Krista~A. Ehinger, Aude Oliva, and Antonio Torralba.
\newblock {SUN} database: Large-scale scene recognition from abbey to zoo.
\newblock In \emph{{CVPR}}, pp.\  3485--3492. {IEEE} Computer Society, 2010.

\bibitem[Xu et~al.(2020)Xu, Zhang, Ni, Li, Wang, Tian, and Zhang]{xu2020adversarial}
Minghao Xu, Jian Zhang, Bingbing Ni, Teng Li, Chengjie Wang, Qi~Tian, and Wenjun Zhang.
\newblock Adversarial domain adaptation with domain mixup.
\newblock In \emph{AAAI}, volume~34, pp.\  6502--6509, 2020.

\bibitem[Yang et~al.(2024)Yang, Zhang, Wang, and Xie]{yang2024mma}
Lingxiao Yang, Ru-Yuan Zhang, Yanchen Wang, and Xiaohua Xie.
\newblock Mma: Multi-modal adapter for vision-language models.
\newblock In \emph{CVPR}, pp.\  23826--23837, 2024.

\bibitem[Yang et~al.(2023)Yang, Panagopoulou, Zhou, Jin, Callison{-}Burch, and Yatskar]{yang2023language}
Yue Yang, Artemis Panagopoulou, Shenghao Zhou, Daniel Jin, Chris Callison{-}Burch, and Mark Yatskar.
\newblock Language in a bottle: Language model guided concept bottlenecks for interpretable image classification.
\newblock In \emph{{CVPR}}, pp.\  19187--19197. {IEEE}, 2023.

\bibitem[Yao et~al.(2023)Yao, Zhang, and Xu]{yao2023visual}
Hantao Yao, Rui Zhang, and Changsheng Xu.
\newblock Visual-language prompt tuning with knowledge-guided context optimization.
\newblock In \emph{CVPR}, pp.\  6757--6767, 2023.

\bibitem[Yao et~al.(2024)Yao, Zhang, and Xu]{yao2023tcp}
Hantao Yao, Rui Zhang, and Changsheng Xu.
\newblock Tcp: Textual-based class-aware prompt tuning for visual-language model.
\newblock In \emph{CVPR}, 2024.

\bibitem[Yao et~al.(2022)Yao, Wang, Li, Zhang, Liang, Zou, and Finn]{yao2022improving}
Huaxiu Yao, Yu~Wang, Sai Li, Linjun Zhang, Weixin Liang, James Zou, and Chelsea Finn.
\newblock Improving out-of-distribution robustness via selective augmentation.
\newblock In \emph{ICML}, pp.\  25407--25437. PMLR, 2022.

\bibitem[You et~al.(2024)You, Mint, Dai, Sekhon, Staib, and Duncan]{you2024calibrating}
Chenyu You, Yifei Mint, Weicheng Dai, Jasjeet~S Sekhon, Lawrence Staib, and James~S Duncan.
\newblock Calibrating multi-modal representations: A pursuit of group robustness without annotations.
\newblock In \emph{CVPR}, pp.\  26140--26150. IEEE, 2024.

\bibitem[Zhai et~al.(2023)Zhai, Mustafa, Kolesnikov, and Beyer]{zhai2023sigmoid}
Xiaohua Zhai, Basil Mustafa, Alexander Kolesnikov, and Lucas Beyer.
\newblock Sigmoid loss for language image pre-training.
\newblock In \emph{ICCV}, pp.\  11975--11986, 2023.

\bibitem[Zhang et~al.(2024{\natexlab{a}})Zhang, Huang, Jin, and Lu]{zhang2024vision}
Jingyi Zhang, Jiaxing Huang, Sheng Jin, and Shijian Lu.
\newblock Vision-language models for vision tasks: A survey.
\newblock \emph{TPAMI}, 2024{\natexlab{a}}.

\bibitem[Zhang \& R{\'e}(2022)Zhang and R{\'e}]{zhang2022contrastive}
Michael Zhang and Christopher R{\'e}.
\newblock Contrastive adapters for foundation model group robustness.
\newblock In \emph{NeurIPS}, volume~35, pp.\  21682--21697, 2022.

\bibitem[Zhang et~al.(2022{\natexlab{a}})Zhang, Sohoni, Zhang, Finn, and R{\'e}]{zhang2022correct}
Michael Zhang, Nimit~S Sohoni, Hongyang~R Zhang, Chelsea Finn, and Christopher R{\'e}.
\newblock Correct-n-contrast: A contrastive approach for improving robustness to spurious correlations.
\newblock In \emph{ICML}, 2022{\natexlab{a}}.

\bibitem[Zhang et~al.(2022{\natexlab{b}})Zhang, Fang, Zhang, Gao, Li, Dai, Qiao, and Li]{zhang2021tip}
Renrui Zhang, Rongyao Fang, Wei Zhang, Peng Gao, Kunchang Li, Jifeng Dai, Yu~Qiao, and Hongsheng Li.
\newblock Tip-adapter: Training-free clip-adapter for better vision-language modeling.
\newblock In \emph{ECCV}, 2022{\natexlab{b}}.

\bibitem[Zhang et~al.(2024{\natexlab{b}})Zhang, Zhu, Tang, Ma, Zhou, and Zhang]{zhang2024dual}
Yabin Zhang, Wenjie Zhu, Hui Tang, Zhiyuan Ma, Kaiyang Zhou, and Lei Zhang.
\newblock Dual memory networks: A versatile adaptation approach for vision-language models.
\newblock In \emph{CVPR}, pp.\  28718--28728, 2024{\natexlab{b}}.

\bibitem[Zhang et~al.(2024{\natexlab{c}})Zhang, Zhang, Yu, Tang, and He]{zhang2024concept}
Yi~Zhang, Ce~Zhang, Ke~Yu, Yushun Tang, and Zhihai He.
\newblock Concept-guided prompt learning for generalization in vision-language models.
\newblock In \emph{AAAI}, 2024{\natexlab{c}}.

\bibitem[Zhou et~al.(2022{\natexlab{a}})Zhou, Yang, Loy, and Liu]{Zhou_2022}
Kaiyang Zhou, Jingkang Yang, Chen~Change Loy, and Ziwei Liu.
\newblock Learning to prompt for vision-language models.
\newblock \emph{IJCV}, 130\penalty0 (9):\penalty0 2337--2348, 2022{\natexlab{a}}.

\bibitem[Zhou et~al.(2022{\natexlab{b}})Zhou, Yang, Loy, and Liu]{zhou2022conditional}
Kaiyang Zhou, Jingkang Yang, Chen~Change Loy, and Ziwei Liu.
\newblock Conditional prompt learning for vision-language models.
\newblock In \emph{{CVPR}}, pp.\  16795--16804. {IEEE}, 2022{\natexlab{b}}.

\end{thebibliography}
\bibliographystyle{iclr2025_conference}
\clearpage
\setcounter{secnumdepth}{0}
\setcounter{tocdepth}{4}
\renewcommand{\baselinestretch}{0.90}\normalsize
{\sc \tableofcontents}
\renewcommand{\baselinestretch}{1.0}\normalsize
\clearpage
\section{A \quad Implementation Details}
\label{sec:A}
\subsection{A.1 \quad Finding Spurious Attributes}
\label{subsec:A1}
We delve into our manual identification process for spurious attributes as described in Section 3.2. Following the approach outlined in~\citep{singla2021salient}, we present a simplified version. For each category, we randomly select 5 images from the training set and generate the corresponding heatmap. We also reference external sources like Wikipedia and seek advice from ChatGPT. Using this information, we assess whether an attribute belongs to the main object or a separate background element, with options: "Yes", "No", or "Unsure". Finally, attributes categorized as "No" are deemed spurious attributes. It is important to mention that unlike~\citep{singla2021salient}, we do not conduct crowd studies. All supervision tasks are performed by the authors.
\subsection{A.2 \quad Prompting LLMs} 
\label{subsec:A2}
We conduct a naive attempt to modify the prompting technique of LLMs to avoid generating spurious attributes in Section 3.2. We try three variant prompt templates by appending or inserting additional instructions as follows: 
\begin{center}
\begin{tabular}{l}
    \textbf{T1:} Only focus on \_\_\_ itself.  \\
    \textbf{T2:} Imagine you are an expert of \_\_\_. \\
    \textbf{T3:} Do not describe other than \_\_\_. \\
  \end{tabular}
\end{center} 
For each instruction, we position it at either the beginning or the end, yielding six combinations. Then, we employ existing attribute-based methods, \eg, ArGue~\citep{tian2023argue}, to derive results, averaging them across all combinations.
\subsection{A.3 \quad Querying MLLMs}
\label{subsec:A3}
In addition to the techniques and parameters introduced in the main paper, we believe a crucial step in dealing with MLLMs is managing their outputs. Given a specified temperature, the output variance of an MLLM, particularly GPT-4V, for the same input can be significant. The responses may range from a single word to a complete paragraph, and the model may fail to follow the demonstrated formats or refuse to respond. Similar challenges have been noted in recent studies, such as DCLIP~\citep{menon2022visual} and CuPL~\citep{pratt2023does}, when using MLLMs or LLMs to generate attributes. In this work, we employ a simple regular expression to retain responses of suitable length and exclude those that are not formatted with bullet points. Additionally, we filter out duplicate or similar attributes. For example, between ${\rm ice \ surface}$ and ${\rm glacier}$ we typically randomly select only one.
\subsection{A.4 \quad Constructing Pseudo Categories}
\label{subsec:A4}
Here, we describe the process of constructing images targeting spurious attributes using SD~\citep{rombach2022high} or LAION-5B~\citep{schuhmann2022laion}. For the former, following Sus-X~\citep{udandarao2022sus}, we use the common checkpoint stable-diffusion-v1-4, with a guidance scale of 7.0. The diffusion step is set to 100, with a fixed output resolution of 512x512. Additionally, to ensure the diversity of the images, we use ChatGPT to generate multiple SD prompts. Specifically, we provide a vanilla prompt as an example, e.g., ${\rm a \ photo \ of \ a \ mouse}$, and then ask GPT to rephrase the prompt in different formats. Some example generated prompts are displayed below.
\begin{center}
\begin{tabular}{l}
    \textbf{P1:} a 3D realistic photo of a \_\_\_  \\
    \textbf{P2:} a high-quality natural image of \_\_\_. \\
    \textbf{P3:} a intriguing portray of \_\_\_. \\
  \end{tabular}
\end{center} 
It is worth noting that the prompts mentioned above are also applicable for pre-training retrieval. For LAION-5B, we select the matches with the highest average semantic similarity to these GPT-generated prompts to construct pseudo categories. This approach ensures the diversity of the retrieved images while also enhancing the reliability of semantic matching.
\section{B \quad More Evaluation}
\label{sec:B}
\subsection{B.1 \quad Example Spurious Attributes} 
\label{subsec:B1}
\texttt{SAP} quantifies the identification of spurious attributes without human supervision, offering a more precise measure of spurious correlation. This aids in effectively pinpointing attributes favored by VLMs. Table~\ref{tab:example_spurious} showcases typical spurious attributes found by \texttt{SAP}, including instances like ${\rm mouse}$ frequently appearing with ${\rm laptop}$, or ${\rm fork}$ being closely associated with ${\rm apple \ pie}$. Additionally, we assess their weights on model predictions, along with the average weights of all generated attributes for reference. Notably, spurious attributes carry substantially higher weights in model decision-making compared to overall attributes, further underscoring the biased nature of VLMs.

\subsection{B.2 \quad SAP vs Human Supervision}
\label{subsec:B2}
\begin{figure*}[t!] 
    \centering
    \includegraphics[width=1\linewidth]{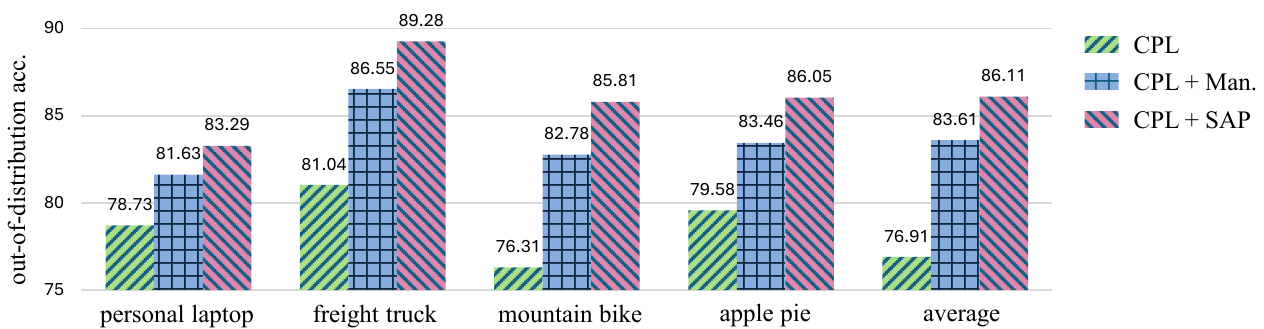}
    \caption{\textbf{The per-category out-of-distribution accuracy on domain generalization.} In this setting, based on the strong baseline CPL, we remove spurious attributes identified by manual inspection (Man.) or \texttt{SAP} for a specific category and compare the accuracy change on the category in the out-of-distribution datasets. All results are averaged over 4 ImageNet variants. 
    }
\label{fig:sap_man}
\end{figure*}

Finding spurious attributes through human supervision~\citep{singla2021salient, wong2021leveraging}, while comprehensive, has significant drawbacks: 1) it incurs extremely high labor costs; 2) its strong subjectivity easily introduces false positives, where identified attributes are only present by chance. Here, we compare the performance of the proposed automatic identification method, \texttt{SAP}, with human supervision. We adopt domain generalization as the task and select CPL~\citep{zhang2024concept} as the baseline. For better interpretation, we remove spurious attributes from individual categories one at a time and record the change in per-category accuracy on out-of-distribution datasets. Fig.~\ref{fig:sap_man} depicts the results in CPL, as well as the results after removing spurious attributes through the two identification approaches. It can be seen that \texttt{SAP}'s performance is comparable or even outperforms human supervision.

\begin{table*}[t]
\footnotesize
\centering
\setlength{\tabcolsep}{6mm}
\renewcommand{\arraystretch}{1.3}
\begin{tabular}{lll}
\ChangeRT{1.2pt}
Category name   & Spurious attributes                  & Average weights \\ \hline
Personal laptop & \textit{mouse, coffee, charger, worktable}    & 77.34\% / 46.73\%      \\
Freight truck   & \textit{road, traffic light, trees, street}   & 82.16\% / 54.69\%     \\
Mountain bike   & \textit{trees, road, mountain, swamp}         & 74.81\% / 43.02\%     \\
Apple pie      & \textit{fork, plates, dining car, tablecloth} & 67.29\% / 37.44\%     \\ \ChangeRT{1.2pt}
\end{tabular}
\caption{\textbf{The spurious attributes identified by \texttt{SAP}.} For each example category, we pinpoint its spurious attributes and determine the average attribute weights on model predictions using CBMs across identified spurious attributes (\textbf{Left}) and all generated attributes (\textbf{Right}).}
\label{tab:example_spurious}
\end{table*}
\subsection{B.3 \quad Querying with SAP at Scale}
\label{subsec:B3}
\begin{table*}[t]
\footnotesize
\centering
\setlength{\tabcolsep}{1.0mm}
\renewcommand{\arraystretch}{1.4}
\begin{tabular}{lccccccc}
\ChangeRT{1.2pt}
\# Query Images & 1     & 2     & 4     & 8     & 16    & 32    & 64    \\ \hline
CoCoOp          & 75.62 & 76.06 & 76.84 & 77.56 & 78.10 & 78.24 & 78.28 \\
MaPLe           & 81.98 & 82.55 & 83.47 & 84.01 & 84.49 & 84.60 & 84.53 \\
PromptSRC       & 82.59 & 83.35 & 84.13 & 84.72 & 85.46 & 85.63 & 85.68 \\ \ChangeRT{1.2pt}
\end{tabular}
\caption{\textbf{The evaluation on base-to-new generalization while querying different number of images per-category.} The results are averaged across 11 datasets.}
\label{tab:SAP_many}
\end{table*}
In the main paper, we address a challenging setting, specifically few-shot scenarios where training data is limited. This leads to a pertinent question: is a small number of images truly adequate for \texttt{SAP} to identify spurious attributes within categories? In other words, would querying more images further enhance \texttt{SAP}'s performance? To investigate the potential of scaling up, we expand the training dataset from 16-shot to 256-shot and have GPT-4V query 1, 2, 4, 8, 16, 32, and 64 randomly selected images from the training shots. We evaluate the average new category accuracy on base-to-new generalization tasks across 11 datasets, comparing three typical baselines: CoCoOp, MaPLe, and PromptSRC. As shown in Table~\ref{tab:SAP_many}, despite the availability of additional shots during training, the results tend to plateau when querying with 16 images. This indicates that even with an expanded training dataset, MLLMs require only around 16 query images to capture sufficient and effective spurious attributes.
\subsection{B.4 \quad Effect of Choices of MLLMs}
\label{subsec:B4}
\begin{table*}[t]
\footnotesize
\centering
\setlength{\tabcolsep}{1.0mm}
\renewcommand{\arraystretch}{1.4}
\begin{tabular}{lcccc}
\ChangeRT{1.2pt}
\multicolumn{1}{c}{MLLM} & BLIP-2 & LLaVA & InternVL & GPT-4V         \\ \hline
CoCoOp                   & 72.28  & 72.79 & 73.14    & \textbf{73.50} \\
MaPLe                    & 76.20  & 76.83 & 77.25    & \textbf{77.69} \\
PromptSRC                & 76.43  & 77.01 & 77.37    & \textbf{77.88} \\ \ChangeRT{1.2pt}
\end{tabular}
\caption{\textbf{The evaluation on base-to-new generalization with various MLLMs.}}
\label{tab:MLLM}
\end{table*}
In previous experiments, we default our MLLM to GPT-4V. Here, we attempt to use more open-sourced MLLMs to comprehensively evaluate the robustness of our proposed method. We consider three recently popular MLLMs: BLIP-2~\citep{li2023blip}, LLaVA~\citep{liu2024visual}, and InternVL~\citep{chen2024internvl}. Table~\ref{tab:MLLM} presents the performance of these different MLLMs on base-to-new generalization tasks with 16-shot. As expected, GPT-4V, the proprietary model, achieves the best results. The next best performance is from InternVL. Conversely, BLIP-2 shows the poorest performance, which we attribute to its tendency to produce a limited vocabulary that results in overly broad core and spurious attributes.
\subsection{B.5 \quad Synthetic Generation vs Pre-training Retrieval}
\label{subsec:B5}
\begin{figure*}[b] 
    \centering
    \includegraphics[width=0.45\linewidth]{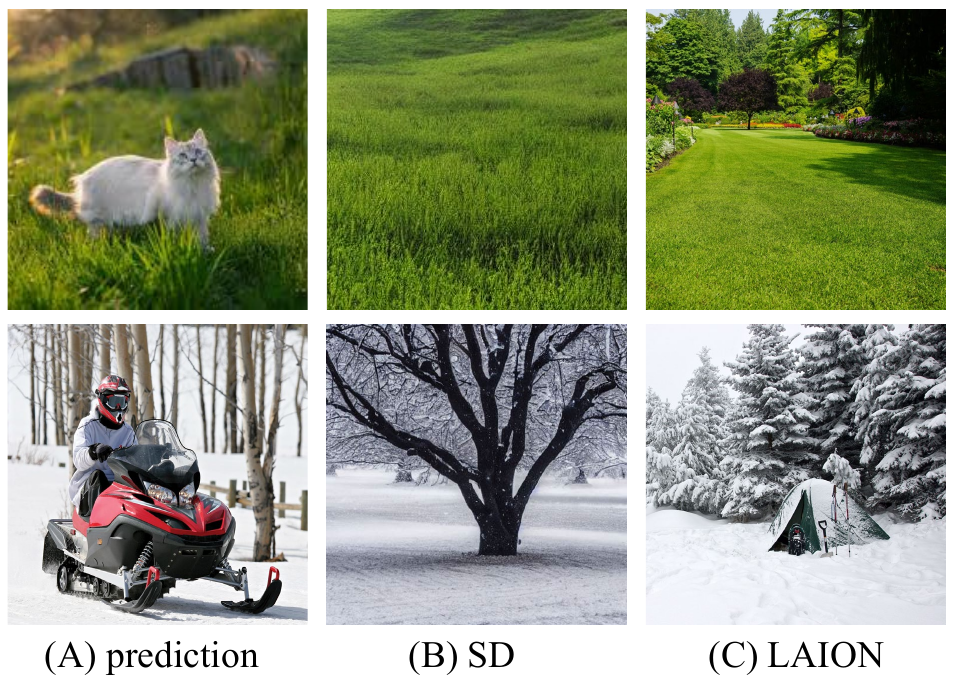}
    \caption{\textbf{Constructed images.}}
\label{fig:constructed}
\end{figure*}
\begin{table*}[t]
\footnotesize
\centering
\setlength{\tabcolsep}{1.0mm}
\renewcommand{\arraystretch}{1.4}
\begin{tabular}{cccccc}
\ChangeRT{1.2pt}
      & CoCoOp         & KgCoOp         & MaPLe          & PromptSRC      & CPL            \\ \hline
SD    & \textbf{76.93} & \textbf{78.51} & \textbf{80.06} & 80.57 & \textbf{82.87} \\
LAION & 76.46          & 78.45          & 79.32          & \textbf{80.85}          & 82.35          \\ \ChangeRT{1.2pt}
\end{tabular}
\caption{\textbf{The comparison between synthetic generation and pre-training dataset retrieval.} We select 5 strong baselines. The results are harmonic mean of accuracy on base and new categories for base-to-new generalization.}
\label{tab:sd_laion}
\end{table*}
In previous experiments, our default approach is to utilize Stable Diffusion for constructing pseudo categories. Here, we explore an alternative method: retrieving image samples from the pre-training dataset. Table~\ref{tab:sd_laion} illustrates the results of both approaches across several baselines. Notably, Stable Diffusion consistently outperforms retrieval from LAION-5B. This unexpected result is intriguing, considering that pre-training images predominantly consist of real data, which one would expect to better match the style of the target category. However, the results suggest otherwise. We speculate that the complexity of real image distributions, coupled with noise attributes, may contribute to this disparity. For instance, in Fig.~\ref{fig:constructed}, when associating the spurious attribute ${\rm snow forest}$ with ${\rm snowmobile}$, the top-1 match retrieved using LAION-5B includes elements such as ${\rm tent}$ and ${\rm bag}$. These noise attributes could potentially introduce new shortcuts, complicating the model's ability to differentiate spurious attributes from the target category.
\subsection{B.6 \quad Balancing the Effect of SAS}
\label{subsec:B6}
\begin{table*}
\footnotesize
\centering
\setlength{\tabcolsep}{1.0mm}
\renewcommand{\arraystretch}{1.4}
\begin{tabular}{ccccccc}
\ChangeRT{1.2pt}
$\lambda$ & 0     & 1     & 2     & 5     & 10    & 20    \\ \hline
Base   & 83.23 & \textbf{83.73} & 83.64 & 83.15 & 82.41 & 81.53 \\
New    & 74.82 & 76.00 & 77.36 & \textbf{77.73} & 76.60 & 76.89 \\
HM     & 78.80 & 79.68 & \textbf{80.38} & 80.35 & 79.40 & 79.14 \\ \ChangeRT{1.2pt}
\end{tabular}
\caption{\textbf{The effect of $\mathcal{L}_{pse}$ with different $\lambda$ on base-to-new generalization.}} 
\label{tab:lambda}
\end{table*}
Table~\ref{tab:lambda} examines the balancing effect between $\mathcal{L}_{pse}$ and primary learning objectives in existing work in terms of $\lambda$. The best trade-off is observed at around $\lambda = 2$. As $\lambda$ increases further, it begins to neglect the primary objectives of the baselines, leading to a decline in base accuracy. Notably, these results are averaged across multiple baselines. In fact, for distinct baselines, we suggest exploring optimal values individually due to their respective learning characteristics.
\subsection{B.7 \quad Quantitative Comparison with Related Work}
\label{subsec:B7}
\begin{table*}[b]
\footnotesize
\centering
\setlength{\tabcolsep}{1.4mm}
\setlength\heavyrulewidth{0.25ex}
\renewcommand{\arraystretch}{1.0}
\begin{tabular*}{\textwidth}{@{}l@{\extracolsep{\fill}}*{12}{c}@{}}
\toprule
Method       & \multicolumn{3}{c}{Waterbirds}                      & \multicolumn{3}{c}{CelebA}                          & \multicolumn{3}{c}{BREEDS}                          & \multicolumn{3}{c}{CIFAR-10.02}                     \\ \cmidrule(){2-4} \cmidrule(){5-7} \cmidrule(){8-10} \cmidrule(){11-13}
Accuracy (\%)     & WG            & Avg  & \cellcolor[HTML]{9AFF99}Gap  & WG            & Avg  & \cellcolor[HTML]{9AFF99}Gap  & WG            & Avg  & \cellcolor[HTML]{9AFF99}Gap  & WG            & Avg  & \cellcolor[HTML]{9AFF99}Gap  \\ \midrule
Zero-shot    & 25.7          & 87.3 & \cellcolor[HTML]{9AFF99}61.6 & 62.1          & 71.9 & \cellcolor[HTML]{9AFF99}9.8  & 4.0           & 86.6 & \cellcolor[HTML]{9AFF99}82.6 & 72.0          & 93.2 & \cellcolor[HTML]{9AFF99}21.2 \\
RoboShot     & 45.2          & 79.9 & \cellcolor[HTML]{9AFF99}34.7 & 82.6          & 85.5 & \cellcolor[HTML]{9AFF99}2.9  & 56.4          & 80.3 & \cellcolor[HTML]{9AFF99}23.9 & 79.1          & 95.6 & \cellcolor[HTML]{9AFF99}16.5 \\ \midrule
Linear Probe & 65.9          & 97.6 & \cellcolor[HTML]{9AFF99}31.7 & 28.3          & 94.7 & \cellcolor[HTML]{9AFF99}66.4 & 84.0          & 98.6 & \cellcolor[HTML]{9AFF99}14.6 & 87.5          & 96.1 & \cellcolor[HTML]{9AFF99}8.6  \\
C-Adapter    & 86.9          & 96.2 & \cellcolor[HTML]{9AFF99}9.3  & 84.6          & 90.4 & \cellcolor[HTML]{9AFF99}5.8  & 80.0          & 97.5 & \cellcolor[HTML]{9AFF99}17.5 & 82.2          & 96.1 & \cellcolor[HTML]{9AFF99}13.9 \\
DISC         & 88.7          & 93.8 & \cellcolor[HTML]{9AFF99}5.1  & 82.0          & 92.5 & \cellcolor[HTML]{9AFF99}10.5 & 86.3          & 95.8 & \cellcolor[HTML]{9AFF99}9.5  & 84.7          & 94.3 & \cellcolor[HTML]{9AFF99}9.6  \\
CFR          & 88.2          & 96.7 & \cellcolor[HTML]{9AFF99}8.5  & 84.7          & 87.8 & \cellcolor[HTML]{9AFF99}3.1  & 85.0          & 96.1 & \cellcolor[HTML]{9AFF99}11.1 & \textbf{89.1} & 92.5 & \cellcolor[HTML]{9AFF99}3.4  \\
\texttt{SAS}          & \textbf{89.7} & 96.3 & \cellcolor[HTML]{9AFF99}6.6  & \textbf{87.4} & 91.1 & \cellcolor[HTML]{9AFF99}3.7  & \textbf{87.8} & 96.4 & \cellcolor[HTML]{9AFF99}8.6  & 88.5          & 95.2 & \cellcolor[HTML]{9AFF99}6.7  \\ \bottomrule
\end{tabular*}
\caption{\textbf{The group robustness evaluation of \texttt{SAS} and other spurious correlation mitigation methods.} We report worst-group accuracy (WG), average-group accuracy (Avg) and the gap between. Note that RoboShot is a zero-shot calibration method, while other approaches are training-required.}
\label{tab:related_work}
\end{table*}
In the main text, we primarily demonstrate the effectiveness of \texttt{SAS} in complementing existing PEFT methods. Here, we further substantiate the advantages of \texttt{SAS} by comparing it with other state-of-the-art spurious correlation mitigation approaches. We evaluate a typical property of VLMs, group robustness, which indicates the invariance of VLMs under different associations between labels and attributes. For the baselines, we consider C-Adapter~\citep{zhang2022contrastive} and CFR~\citep{you2024calibrating}, where spurious attributes are assumed to be unknown. We also include RoboShot~\citep{adila2023zero} and DISC~\citep{wu2023discover}, where, similar to our approach, spurious concepts are identified and used for precise mitigation. By default, we configure \texttt{SAS} to optimize only the learnable textual prompt, \ie, CoOp~\citep{Zhou_2022}. It is worth noting that RoboShot~\citep{adila2023zero} is a zero-shot approach that calibrates pre-trained embeddings. Following \citet{zhang2022contrastive}, we consider four datasets with group annotations: Waterbirds~\citep{sagawa2019distributionally}, CelebA~\citep{liu2018large}, BREEDS Living-17~\citep{santurkar2020breeds}, and CIFAR-10.02~\citep{krizhevsky2009learning}. In Table~\ref{tab:related_work}, average-group accuracy, worst-group accuracy, and their gap are reported. It can be observed that \texttt{SAS} achieves a new state-of-the-art in worst-group accuracy across most datasets without excessively compromising average-group accuracy.
\subsection{B.8 \quad Generalization under Limited Shots}
\label{subsec:B8}
\begin{figure*}[t] 
    \centering
    \includegraphics[width=0.6\linewidth]{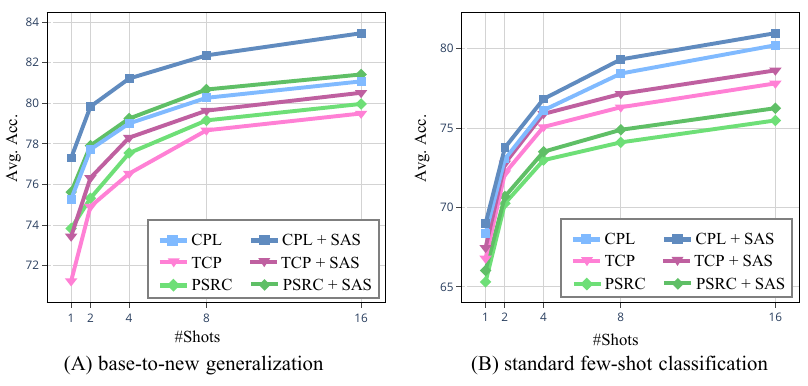}
    \caption{\textbf{The results varying shots on base-to-new generalization and few-shot classifcation.}}
\label{fig:fewshot}
\end{figure*}
We consider generalization capability in extreme cases, where the shots are further limited, \ie, 1/2/4/8 shots. It is noteworthy that in this scenario, limitations arise from both the insufficient amount of data and the impact on \texttt{SAP}'s precision to identify spurious attributes, further affecting \texttt{SAS} performance. We select three strong baselines, encompassing PromptSRC~\citep{khattak2023self}, TCP~\citep{yao2023tcp} and CPL~\citep{zhang2024concept}. Fig.~\ref{fig:fewshot} (A) shows the results of combining \texttt{SAS} on base-to-new generalization across different shot settings. It can be seen that the results consistently outperform the original baselines, even only one shot is given.
\subsection{B.9 \quad Standard Few-shot Classification}
\label{subsec:B9}
We consider the standard scenario where test and training samples originate from the same dataset distribution. Fig.~\ref{fig:fewshot} (B) illustrates the results in standard few-shot classification. Notably, integrating \texttt{SAS} does not compromise in-distribution accuracy; instead, it shows a slight and consistent improvement.
\subsection{B.10 \quad Discussion of Hyperparameter Sensitivity}
\label{subsec:B10}
Although we observe the state-of-the-art performance of \texttt{SAS} in the main context, an important aspect, hyperparameter sensitivity, still requires discussion. For the newly introduced hyperparameters in \texttt{SAS}, such as $\lambda$ and $\gamma$, their impact on the results has been examined in previous ablation experiments. These experiments reveal that while an optimal value is preferred, \texttt{SAS} is not overly sensitive to these hyperparameters and consistently provides stable improvements within a certain range. 

Regarding training hyperparameters such as learning rate and batch size, recent studies~\citep{silva2024closer} have found that some adaptation methods heavily rely on these hyperparameters in few-shot scenarios, complicating practical deployment. In contrast, as shown in Fig. 3 of the main paper, although \texttt{SAS} uses different training hyperparameters for different baselines as specified in the original papers, it consistently achieves gains, demonstrating its robustness to hyperparameters.
\subsection{B.11 \quad Ablation on Performance Gains}
{In the main paper, we verify the effectiveness of \texttt{SAS} and explore the contribution of spurious attributes to its performance. To further confirm that the performance gains are primarily due to the model's enhanced robustness to spurious attributes rather than additional data, here we conduct a simple ablation study. Specifically, in addition to the proposed method, we design two baselines. In the first baseline, we consider additional data directly from the original dataset featuring the main objects, where we extend the training data from 16 shots to 32 shots (32-shot main).  In the second baseline, we involve additional data generated by pseudo categories, where instead of featuring spurious attributes, these pseudo categories are the same as the main categories, i.e., vanilla constructed data (16-shot main + 16-shot pseudo main). In contrast to the first two baselines, our approach creates pseudo categories based on spurious attributes (16-shot main + 16-shot pseudo spurious). For fairness, we ensure that the amount of training data is identical between the two baselines and our approach. We select three typical methods for comparison, including CoCoOp~\citep{zhou2022conditional}, MaPLe~\citep{MaPLe}, and PromptSRC~\citep{khattak2023self}, and evaluate them on the base-to-new generalization task. All results are averaged across 11 datasets.

As shown in Table~\ref{tab:performance_gains}, generating additional data using spurious attributes significantly outperforms vanilla constructed data for main categories ($76.36\%$ vs $73.53\%$). Furthermore, our proposed method even exceeds the performance of the 32-shot main ($76.36\%$ vs $74.34\%$). It is important to note that this comparison is not entirely fair for our method, as the latter relies on more labeled data from the original training set. This further suggests that the performance gains are primarily driven by the model's enhanced robustness to spurious attributes, rather than merely the increased training data. 
}
\subsection{B.12 \quad More Visualization Examples}
\label{subsec:B12}
\begin{table*}[t]
\footnotesize
\centering
\setlength{\tabcolsep}{1.2mm}
\renewcommand{\arraystretch}{1.3}
\begin{tabular}{lcccc}
\ChangeRT{1.2pt}
\multicolumn{1}{c}{Training Data}             & CoCoOp    & MaPLe     & PromptSRC & Average   \\ \hline
16-shot main                                  & 70.05     & 75.39     & 73.78     & 72.94     \\
\hline
32-shot main                                  & 71.12     & 76.37     & 75.52     & 74.34     \\
16-shot main + 16-shot pseudo main            & 70.34     & 75.80     & 74.46     & 73.53     \\
16-shot main + 16-shot pseudo spurious (ours) & \textbf{73.50} & \textbf{77.69} & \textbf{77.88} & \textbf{76.36} \\ \ChangeRT{1.2pt}
\end{tabular}
\caption{\textbf{The ablation study on the performance gains}. We introduce two baselines, where the first incorporates additional data from the training set (32-shot main), and the second involves vanilla constructed data from pseudo categories mirroring the main categories (16-shot main + 16-shot pseudo main). In contrast, the pseudo categories of our method feature spurious attributes (16-shot main + 16-shot pseudo spurious)}
\label{tab:performance_gains}
\end{table*}
\begin{figure*}[b] 
    \centering
    \includegraphics[width=1.0\linewidth]{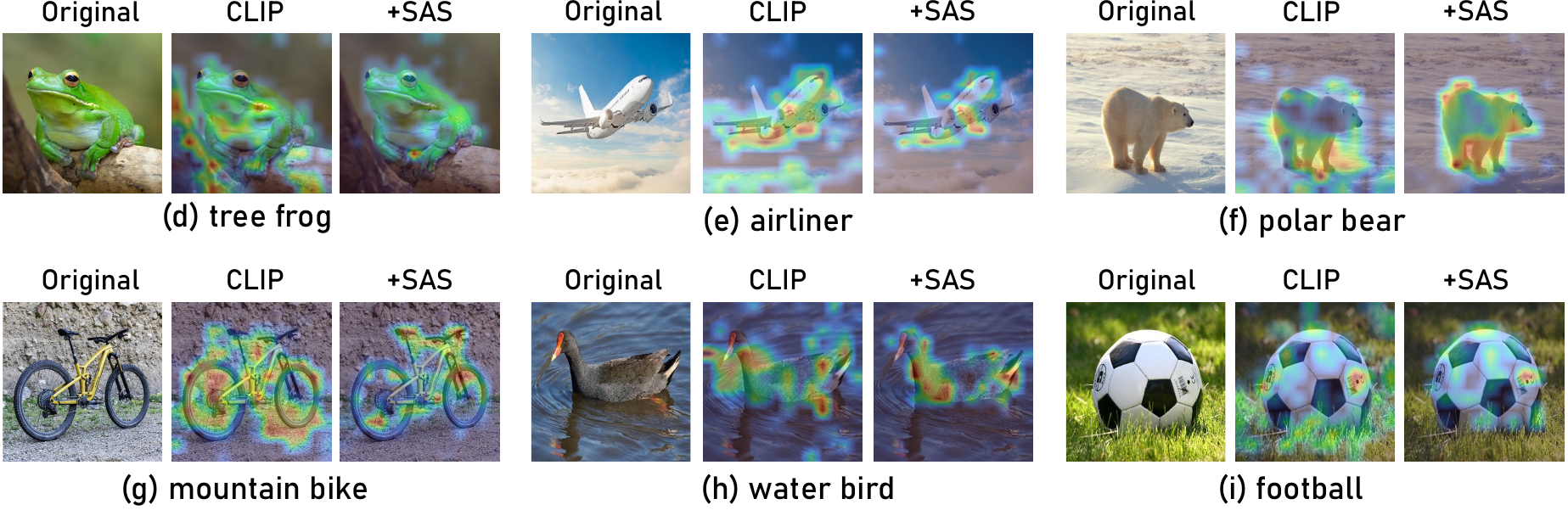}
    \caption{\textbf{More saliency map visualization with and without \texttt{SAS}}. }
\label{fig:more_saliency}
\end{figure*}
{In Fig.~\ref{fig:saliency}, we present the saliency maps for some typical categories with and without \texttt{SAS}. For completeness, we provide more examples here. As shown in Fig.~\ref{fig:more_saliency}, \texttt{SAS} consistently reduces VLMs' bias towards spurious cues across various categories, enabling a greater focus on the main objects. For example, for the tree frog, \texttt{SAS} reduces VLMs' reliance on tree branches, while for the airliner, the typical spurious attributes are sky or clouds, and the application of \texttt{SAS} alleviates the model's bias towards these elements.}
\subsection{B.13 \quad More Vision-Language Models}
\label{subsec:B13}
\begin{table*}[t]
\footnotesize
\centering
\setlength{\tabcolsep}{1.2mm}
\renewcommand{\arraystretch}{1.3}
\begin{tabular}{lcccc}
\ChangeRT{1.2pt}
\multicolumn{1}{c}{Model} & CoCoOp         & MaPLe          & PromptSRC      & Average        \\ \hline
BLIP                      & 68.81          & 72.32          & 72.62          & 70.58          \\
BLIP + \texttt{SAS}                & 70.54          & 74.35          & 73.97          & 72.95          \\
\hline
CLIPA-v2                  & 70.28          & 73.40          & 74.52          & 72.73          \\
CLIPA-v2 + \texttt{SAS}            & 72.42          & 74.88          & 77.08          & 74.79          \\
\hline
EVA-CLIP                  & 72.75          & 77.58          & 76.13          & 75.49          \\
EVA-CLIP + \texttt{SAS}            & 74.60          & \textbf{77.92} & 77.82          & 76.78          \\
\hline
SigLIP                    & 74.99          & 73.78          & 78.64          & 75.80          \\
SigLIP + \texttt{SAS}              & \textbf{76.41} & 75.26          & \textbf{79.87} & \textbf{77.18} \\ \ChangeRT{1.2pt}
\end{tabular}
\caption{\textbf{The evaluation results of \texttt{SAS} on other VLMs}. We consider four representative VLMs including BLIP, CLIPA-v2, EVA-CLIP and SigLIP.}
\label{tab:more_vlms}
\end{table*}
{For completeness, here we extend the evaluation of \texttt{SAS} to additional VLMs other than CLIP. Specifically, we select four typical VLMs encompassing BLIP~\citep{li2022blip}, CLIPA-v2~\citep{li2024inverse}, EVA-CLIP~\citep{sun2023eva} and SigLIP~\citep{zhai2023sigmoid}. We record the results on base-to-new generalization, where the setting is consistent with the main paper. As demonstrated in Table~\ref{tab:more_vlms}, our proposed method, \texttt{SAS}, consistently yields performance gains across a range of VLMs, extending beyond just CLIP.
}
\subsection{B.14 \quad Computational Efficiency and Cost}
\label{subsec:B14}
{In this section, we present the computational and time costs of the proposed method, accompanied by a thorough analysis.}

\begin{table*}[t]
\footnotesize
\centering
\setlength{\tabcolsep}{1.2mm}
\renewcommand{\arraystretch}{1.3}
\begin{tabular}{lccccc}
\ChangeRT{1.2pt}
\multicolumn{1}{c}{Method} & Flowers102 & Food101 & FGVCAircraft & StanfordCars & Average Time \\ \hline
CoCoOp                     & 10m48s     & 18m34s  & 8m02s        & 12m46s       & 12m32s       \\
+ \texttt{SAS}                      & 13m14s     & 24m07s  & 13m36s       & 17m08s       & 17m01s       \\
+ selective trick          & 11m03s     & 20m45s  & 10m23s       & 14m55s       & 14m16s       \\
\hline
PromptSRC                  & 6m25s      & 15m09s  & 5m44s        & 9m36s        & 9m13s        \\
+ \texttt{SAS}                      & 8m26s      & 18m21s  & 7m11s        & 12m04s       & 11m30s       \\
+ selective trick          & 6m58s      & 16m22s  & 6m35s        & 10m20s       & 10m03s       \\ \ChangeRT{1.2pt}
\end{tabular}
\caption{\textbf{The training efficiency of \texttt{SAS} and selective optimization on other datasets}.}
\label{tab:training_time}
\end{table*}
{\textbf{The cost of training}. In the main paper, we present the training time of \texttt{SAS} and the proposed selective optimization trick on ImageNet. Here, furthermore, we provide time statistics for other datasets. The time is measured as the runtime of the training script based on the implementation of CoOp~\citep{Zhou_2022}. As shown in Table~\ref{tab:training_time}, for most datasets, the integration of \texttt{SAS} only increases the training time by approximately $3$ to $5$ minutes, while selective optimization further reduces this time to a negligible amount. In fact, the selective optimization trick is proposed to address large-scale datasets, such as ImageNet, which contains $1000$ categories. For regular datasets ($\sim 100$ categories), the time consumption of \texttt{SAS} is fully acceptable.} 

\begin{table*}[t]
\footnotesize
\centering
\setlength{\tabcolsep}{1.2mm}
\renewcommand{\arraystretch}{1.5}
\begin{tabular}{cccccccclccc}
\ChangeRT{1.2pt}
Dataset        & Caltech & Pets  & Cars  & Flowers & Food  & Aircraft & SUN   & DTD   & EuroSAT & UCF   & INet    \\ \hline
 Time & 25min   & 10min & 45min & 30min   & 25min & 30min    & 90min & 15min & 5min    & 25min & 3h50min \\ \ChangeRT{1.2pt}
\end{tabular}
\caption{\textbf{The diffusion inference time for each dataset}.}
\label{tab:diffusion_time}
\end{table*}
{\textbf{The cost of diffusion generation}. Here, we provide the estimated inference time required to construct pseudo categories through Stable Diffusion for each dataset. As shown in Table~\ref{tab:diffusion_time}, the total inference time is proportional to the size of the dataset, particularly the number of categories involved. For most datasets, the inference time is under half an hour, and the entire inference process can be completed within half a day. It is important to note that this is a one-time operation, and no additional inference is needed during subsequent training.}

\begin{table*}[t]
\footnotesize
\centering
\setlength{\tabcolsep}{1.2mm}
\renewcommand{\arraystretch}{1.5}
\begin{tabular}{cccccccclccc}
\ChangeRT{1.2pt}
Dataset        & Caltech & Pets  & Cars  & Flowers & Food  & Aircraft & SUN   & DTD  & EuroSAT & UCF   & INet    \\ \hline
Time & 10min   & 10min & 25min & 10min   & 10min & 10min    & 35min & 5min & 3min    & 10min & 1h30min \\ \ChangeRT{1.2pt}
\end{tabular}
\caption{\textbf{The GPT prompting time for each dataset}.}
\label{tab:gpt_time}
\end{table*}
{\textbf{The cost of GPT prompting}. In our method, a key step is identifying the spurious attributes within each category, which we accomplish by prompting MLLMs, i.e., GPT. Here we provide the time cost of this process along with a thorough analysis. Specifically, to enhance efficiency, we employ batch inference as implemented in \citet{menon2022visual}, where multiple queries can be processed concurrently, which significantly reduces the inference time for GPT. As shown in Table~\ref{tab:gpt_time}, the GPT inference time for most datasets is under $10$ minutes. The complete inference process takes approximately three hours, which is also a one-time operation that does not need to be repeated thereafter. It is worth noting that upon obtaining the responses, we need to perform post-processing such as filtering and selection to determine valid attributes, as detailed in Section~\hyperref[subsec:A3]{A.3}, which may require additional time. }

\subsection{B.15 \quad More Modalities and Tasks}
\label{subsec:B15}
\begin{table*}[b]
\footnotesize
\centering
\setlength{\tabcolsep}{1.2mm}
\renewcommand{\arraystretch}{1.4}
\begin{tabular}{lllll}
\ChangeRT{1.2pt}
Method          & K-400          & HMDB-51        & UCF-101        & SSv2           \\ \hline
ViFiCLIP        & 61.10          & 53.30          & 67.70          & 12.10          \\ \hline
CoCoOp          & 64.70          & 54.41          & 68.21          & 14.24          \\
CoCoOp + \texttt{SAS}    & 66.39          & 56.64          & 70.40          & 16.01          \\ \hline
MaPLe           & 64.52          & 58.23          & 70.73          & 14.74          \\
MaPLe + \texttt{SAS}     & 66.42          & 59.32          & 72.66          & 16.40          \\ \hline
PromptSRC       & 68.31          & 62.38          & 76.79          & 17.22          \\
PromptSRC + \texttt{SAS} & \textbf{70.23} & \textbf{64.70} & \textbf{79.31} & \textbf{18.95} \\ \ChangeRT{1.2pt}
\end{tabular}
\caption{\textbf{The evaluation results of \texttt{SAS} on four video datasets}. The training is based on ViFiCLIP, a fully fine-tuned CLIP model for video reasoning.}
\label{tab:video}
\end{table*}
{To assess the transferability of our method to other modalities or tasks, we explore video recognition and leave more tasks, such as language reasoning, for future work. Specifically, we choose ViFi-CLIP~\citep{rasheed2023fine}, a fully fine-tuned CLIP model tailored for video understanding. ViFi-CLIP employs a training framework similar to CLIP, incorporating a temporal pooling layer to derive video representations from multiple frames. Following the base-to-new generalization setting in \citet{rasheed2023fine}, we evaluate video-level generalization performance on four video datasets: K-400~\citep{kay2017kinetics}, HMDB-51~\citep{kuehne2011hmdb}, UCF-101~\citep{soomro2012ucf101}, and SSv2~\citep{goyal2017something}. As in the main paper, we select three representative baseline methods: CoCoOp~\citep{zhou2022conditional}, MaPLe~\citep{MaPLe}, and PromptSRC~\citep{khattak2023self}. Since ViFi-CLIP shares its architecture with CLIP, these methods can be easily transferred to ViFi-CLIP, which has been implemented by \citet{khattak2023self}. We incorporate the proposed method, \texttt{SAS}, into these baselines to verify its effectiveness by contrasting spurious attributes with each frame of the video. We record the new category accuracy for each dataset which directly reflects the generalization performance on unseen categories. As shown in Table~\ref{tab:video}, despite the input modalities shifting from images to videos, \texttt{SAS} consistently delivers performance gains across all datasets, proving it to be an effective plug-and-play method that can be generalized to more complex modalities and tasks.}

\subsection{B.16 \quad Evaluation on More Baselines}
\label{subsec:B16}
\begin{table*}[t]
\footnotesize
\centering
\setlength{\tabcolsep}{1.2mm}
\renewcommand{\arraystretch}{1.4}
\begin{tabular}{lccccc}
\ChangeRT{1.2pt}
Method    & ImageNet       & Flowers102     & SUN397         & FGVCAircraft   & StanfordCars   \\ \hline
MMA       & 71.00          & 75.93          & 78.57          & 36.33          & 73.10          \\
MMA + \texttt{SAS} & 72.61          & 77.27          & \textbf{80.19} & \textbf{37.85} & 75.46          \\ \hline
DMN       & 72.28          & 78.49          & 77.32          & 32.60          & 74.22          \\
DMN + \texttt{SAS} & \textbf{73.34} & \textbf{80.17} & 79.74          & 35.38          & \textbf{76.30} \\ \ChangeRT{1.2pt}
\end{tabular}
\caption{\textbf{The evaluation of \texttt{SAS} on other baselines}. We include two recently proposed approaches, including MMA and DMN.}
\label{tab:more_baselines}
\end{table*}
{For completeness, here we evaluate our method on the two recently proposed works. Specifically, we select MMA~\citep{yang2024mma} and DMN~\citep{zhang2024dual}. For the former, we train the newly introduced adapters in the deep layers that bridge the text and image representations, following their setting and implementation. For the latter, we optimize its memory projection functions and incorporate both the static and dynamic memory networks, which is the strongest variant according to their paper. We select the base-to-new generalization task, as illustrated in Section 4, and record the new category accuracy, which directly reflects the generalization performance. As shown in Table~\ref{tab:more_baselines}, \texttt{SAS} consistently improves performance on both methods, demonstrating its complementarity.}

\subsection{B.17 \quad Ablation on Diffusion Steps}
\label{subsec:B17}
\begin{table*}[t]
\footnotesize
\centering
\setlength{\tabcolsep}{1.2mm}
\renewcommand{\arraystretch}{1.3}
\begin{tabular}{ccccccccccc}
\ChangeRT{1.2pt}
\multicolumn{1}{l}{\multirow{2}{*}{Step}} & \multicolumn{2}{c}{Flowers102} & \multicolumn{2}{c}{Food101} & \multicolumn{2}{c}{FGVCAircraft} & \multicolumn{2}{c}{StanfordCars} & \multicolumn{2}{c}{Average} \\ \cline{2-11} 
\multicolumn{1}{l}{}                      & Acc            & Time          & Acc          & Time         & Acc             & Time           & Acc             & Time           & Acc          & Time         \\ \hline
25                                        & 72.24          & 10min         & 91.18        & 6min         & 27.23           & 8min           & 73.55           & 13min          & 66.05        & 9min         \\ \hline
50                                        & 72.85          & 15min         & 91.96        & 11min        & \textbf{28.41}           & 14min          & 74.67           & 23min          & 66.97        & 16min        \\ \hline
75                                        & 72.81          & 22min         & \textbf{92.28}        & 18min        & 28.19           & 24min          & 74.82           & 31min          & 67.02        & 24min        \\ \hline
100                                       & \textbf{72.99}          & 28min         & 92.12        & 26min        & 28.30           & 31min          & \textbf{74.96}           & 40min          & \textbf{67.09}        & 31min        \\ \ChangeRT{1.2pt}
\end{tabular}
\caption{\textbf{The performance of SAS and diffusion time with different number of diffusion steps.}}
\label{tab:diffusion}
\end{table*}
{Considering the computations introduced by diffusion in generating images, here we perform an ablation study on the efficiency of diffusion inference. Specifically, we vary the number of diffusion steps, which is the key hyperparameter influencing the inference time. Intuitively, fewer steps are more efficient yet yield lower image quality, while more steps ensure image fidelity but require more computation. We select CoCoOp as the baseline and record the new category accuracy on base-to-new generalization. As shown in Table~\ref{tab:diffusion}, by default, we use $100$ steps throughout the paper as described in Section~\hyperref[subsec:A4]{A.4}, which requires an average of $31$ minutes to generate images per dataset. Here we try fewer steps, such as $50$, and observe that the time required for diffusion nearly halves ($31$min $\rightarrow$ $16$min) with minimal degradation in performance ($67.09$ $\rightarrow$ $66.97$). However, while the number of steps is further reduced to 25, there is a dramatic performance drop ($66.97$ $\rightarrow$ $66.05$), possibly due to the decline in image quality. This suggests we may safely adjust the number of steps from $100$ to $50$, which halves the required time with minimal accuracy loss, significantly improving the efficiency of \texttt{SAS}.}

\section{C \quad Further Exploration}
\label{sec:C}
\begin{table*}[t]
\footnotesize
\centering
\setlength{\tabcolsep}{1.0mm}
\renewcommand{\arraystretch}{1.2}
\begin{tabular}{lcccccccc}
\ChangeRT{1.2pt}
\multicolumn{1}{c}{Method} & ImageNet & Caltech101 & OxfordPets & StanfordCars & DTD   & EuroSAT & ImageNet-A & Average \\ \hline
CLIP                       & 66.54    & 94.62      & 90.41      & 64.69        & 44.84 & 47.50   & 49.32      & 65.42   \\ \hline
DCLIP                      & 68.52    & 95.48      & 91.88      & 65.70        & 45.52 & 49.08   & 49.88      & 66.58   \\
DCLIP - SA                 & 67.67    & 94.76      & 91.31      & 64.79        & 45.06 & 47.92   & 49.30      & 65.83   \\ \hline
CuPL                       & 69.99    & 96.51      & 92.62      & 66.91        & 47.32 & 50.33   & 50.14      & 67.69   \\
CuPL - SA                  & 68.70    & 95.89      & 92.38      & 65.66        & 46.06 & 49.58   & 49.28      & 66.79   \\ \ChangeRT{1.2pt}
\end{tabular}
\caption{\textbf{The zero-shot accuracy before and after removing spurious attributes.} The model is evaluated on 2 generic datasets (ImageNet~\citep{deng2009imagenet}, Caltech101~\citep{fei2004learning}), 2 fine-grained datasets (OxfordPets~\citep{parkhi12a}, StanfordCars~\citep{krause20133d}), 2 specialized datasets (DTD~\citep{cimpoi2014describing}, EuroSAT~\citep{helber2019eurosat}) and 1 adversarial dataset (ImageNet-A~\citep{hendrycks2021natural}).}
\label{tab:zero-shot}
\end{table*}
\subsection{C.1 \quad Spurious Attributes for Zero-shot Recognition}
\label{subsec:C1}
The primary takeaway of this paper is the unbalanced treatment of various semantic attributes by VLMs, which extends beyond the generalization task and suggests that the language encoder of VLMs may allocate distinct attention to different tokens. We examine a typical example: zero-shot recognition, where attributes are directly utilized to make predictions without training. We consider three baselines: CLIP~\citep{radford2021learning}, DCLIP~\citep{menon2022visual}, and CuPL~\citep{pratt2023does}. The latter two employ LLMs to generate attributes and enhance zero-shot accuracy. In a manner similar to the previous motivational study, we remove identified spurious attributes from the existing baselines and record the accuracy before and after this intervention. Table~\ref{tab:zero-shot} presents the results before and after removal. We observe a significant drop in accuracy for the baselines (from $63.23\%$ to $62.49\%$ for DCLIP and from $64.34\%$ to $63.45\%$ for CuPL), with DCLIP almost reverting to the performance of vanilla CLIP ($62.49\%$ vs. $62.07\%$). This indicates that 1) similar to the generalization task, zero-shot recognition is also dominated by spurious attributes, nearly ignoring the presence of other generated attributes; and 2) spurious attributes, in a sense, improve zero-shot performance on natural datasets by scaling up the model's inherent bias.
\subsection{C.2 \quad Selective Optimization Trick}
\label{subsec:C2}
\texttt{SAS} introduces a subsidiary task that includes constructed pseudo categories and auxiliary learning objectives. With an increasing number of spurious attributes, a large number of pseudo categories are introduced, significantly increasing computational costs. To tackle this challenge, we introduce a strategy that selectively optimizes partial target categories with a heavy bias towards spurious attributes. In other words, we only mitigate the influence of spurious attributes on categories that overly rely on them. To identify these categories, we propose Spurious Correlation Ratio (SCR). SCR is calculated as the ratio of the average weights of spurious attributes to the average weights of all attributes, as exemplified in the rightmost column of Table~\ref{tab:example_spurious}. A higher SCR indicates that the prediction of the corresponding category relies more on spurious attributes. In implementing this trick, we empirically select only the top 10\% of categories ranked by SCR for optimization. To verify the trick, 
we choose two time-intensive baselines, CoCoOp~\citep{zhou2022conditional} and PromptSRC~\citep{khattak2023self}, for comparison. CoCoOp's training is slow due to its instance-conditioned mechanism, while PromptSRC adds three extra learning objectives to the original cross-entropy loss. To emphasize the results, we conduct evaluation on the base-to-new generalization task using ImageNet~\citep{deng2009imagenet} and record both training time and harmonic mean accuracy. Table~5 in the main paper illustrates the trade-off between effectiveness and efficiency with \texttt{SAS} and the proposed trick. It is evident that integrated with selective optimization, the required time is significantly reduced compared to the original \texttt{SAS}. For instance, on PromptSRC, it only adds 9 minutes of training time while preserving most of the performance gains.
\subsection{C.3 \quad Variants of SAS}
\label{subsec:C3}
\begin{figure*}[t] 
    \centering
    \includegraphics[width=0.45\linewidth]{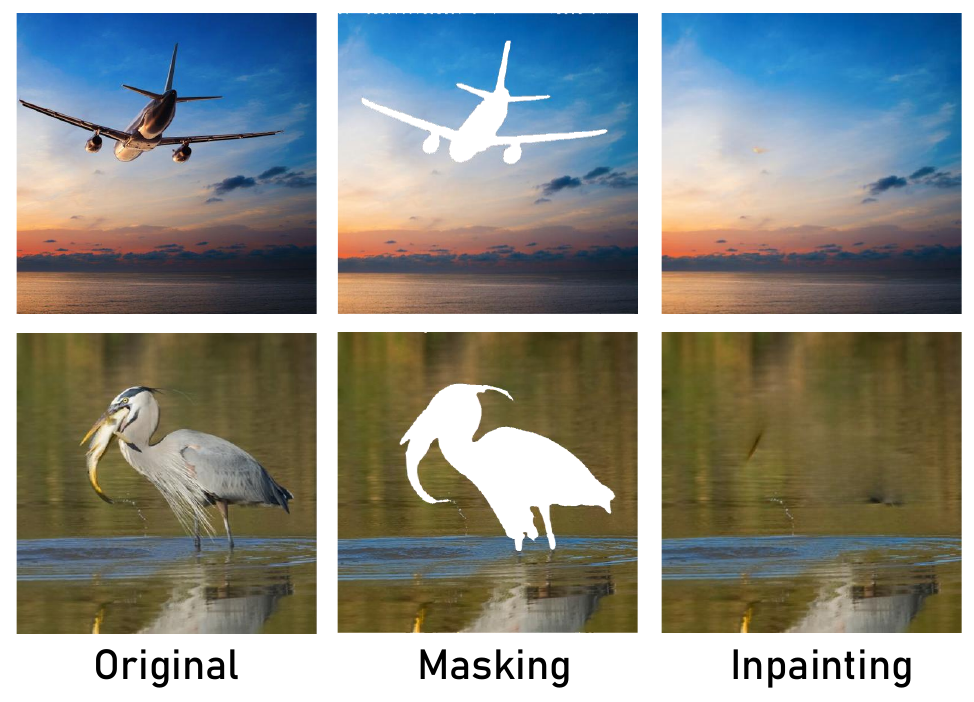}
    \caption{\textbf{The constructed categories with masking and inpainting.
    } For the former, we directly mask the primary object with SAM~\citep{kirillov2023segment}, whereas for the latter, we further use RePaint~\citep{lugmayrinpainting} to fill in the missing parts. We compare the performance between the pseudo categories constructed with masking, inpainting and our synthesis method.}
\label{fig:inpainting}
\end{figure*}
\begin{table*}[b]
\footnotesize
\centering
\setlength{\tabcolsep}{1.0mm}
\renewcommand{\arraystretch}{1.4}
\begin{tabular}{lccc}
\ChangeRT{1.2pt}
\multicolumn{1}{c}{Method}         & CoCoOp & MaPLe & PromptSRC \\ \hline
SAS-masking    & 71.08  & 74.53 & 75.39     \\
SAS-inpainting & 72.95  & 76.81 & 77.04     \\
SAS-synthesis  & \textbf{73.50}  & \textbf{77.69} & \textbf{77.88}     \\ \ChangeRT{1.2pt}
\end{tabular}
\caption{\textbf{The evaluation on base-to-new generalization with two \texttt{SAS} variants.}}
\label{tab:variants}
\end{table*}
In the main paper, \texttt{SAS} primarily constructs pseudo categories using synthetic or pre-trained data, which has proven effective. Here, we consider two simple yet direct variants: 1) instead of utilizing spurious attributes to create new data, we directly mask the main object in the original images and use these as the corresponding pseudo categories, termed \texttt{SAS-masking}; 2) upon masking, we fill the masked area through in-painting, termed \texttt{SAS-inpainting}. Fig.~\ref{fig:inpainting} displays some example images of pseudo categories by these two variants. The motivation here is to enhance VLMs' awareness of core attributes by contrasting target categories with their corresponding images that lack main objects. We refer to the original approach as \texttt{SAS-synthesis}, where pseudo categories are constructed with SD-synthesized images. Table~\ref{tab:variants} presents the performance of the three methods, showing that both variants perform worse than \texttt{SAS-synthesis}. We speculate that this is because 1) masking or in-painting significantly reduces image fidelity; and 2) this approach introduces excessive noise attributes, thereby forming a new set of spurious attributes for VLMs to learn.
\subsection{C.4 \quad Non-semantic Spurious Attribute}
\label{subsec:C4}
\begin{figure*}[b] 
    \centering
    \includegraphics[width=0.4\linewidth]{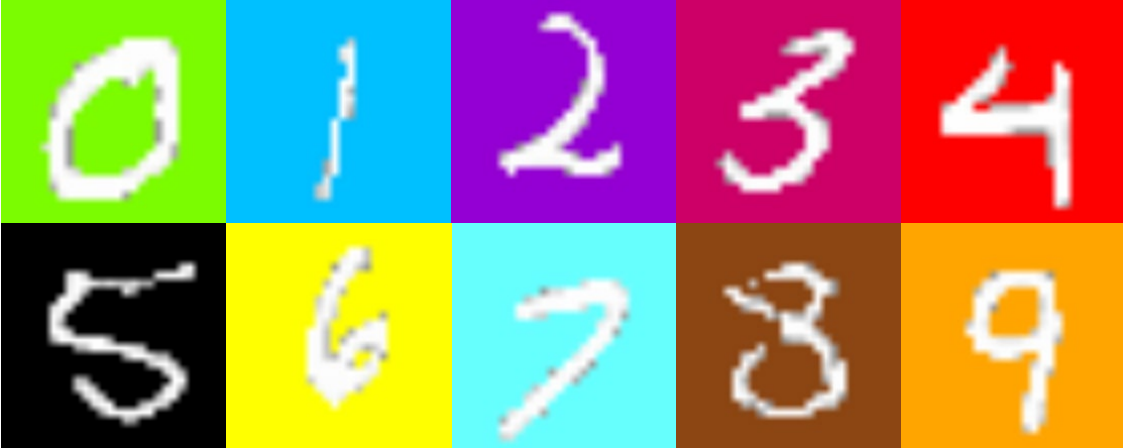}
    \caption{\textbf{ColoredMNIST.
    }}
\label{fig:cmnist}
\end{figure*}
\begin{table*}[t]
\footnotesize
\centering
\setlength{\tabcolsep}{1.0mm}
\renewcommand{\arraystretch}{1.4}
\begin{tabular}{lccc}
\ChangeRT{1.2pt}
\multicolumn{1}{c}{Method} & CoCoOp & MaPLe & PromptSRC \\ \hline
w/o \texttt{SAS}                    & 72.47  & 75.89 & 74.08     \\
w/ \texttt{SAS}                     & 84.70  & 88.27 & 87.48     \\ \ChangeRT{1.2pt}
\end{tabular}
\caption{\textbf{The evaluation on ColoredMNIST with and without \texttt{SAS}.}}
\label{tab:cmnist}
\end{table*}
In the evaluated datasets, including previous work on measuring group robustness~\citep{zhang2022contrastive, adila2023zero}, most spurious attributes are semantically related, wherein the attribute and label exhibit a natural association, \eg, water and water bird. In this study, we extend our exploration to non-semantic attributes, where the association between the attribute and label is artificially constructed. We implement a straightforward color-shifting experiment using ColoredMNIST. This dataset comprises 10 classes, each representing a digit; however, instead of the standard black background in MNIST, each digit class features a distinctly colored background. Each color demonstrates a strong spurious correlation with its corresponding digit, effectively serving as a spurious attribute. Fig.~\ref{fig:cmnist} illustrates examples from ColoredMNIST. We employ GPT-4V to identify these non-semantic spurious attributes, resulting in descriptors such as ${\rm green \ background}$ and ${\rm pure \ yellow \ background}$. We evaluate \texttt{SAS} on the test set of ColoredMNIST, where the color backgrounds are randomized across labels. As shown in Table~\ref{tab:cmnist}, \texttt{SAS} significantly enhances VLMs' robustness to color shifting, indicating that MLLMs may capture non-semantic attributes in images, and \texttt{SAS} effectively leverages these attributes to improve generalization.
\subsection{C.5 \quad Limitation and Failure Cases}
\label{subsec:C5}
In \texttt{SAP}, the primary limitation stems from the necessity of having available images. Previous approaches to generating visual attributes only require textual information, \eg, category names. The underlying assumption is that the generated attributes would be dataset-agnostic. For example, attributes like ${\rm headlights}$, ${\rm doors}$, or ${\rm wheels}$ for the category ${\rm vehicle}$ are assumed to be consistent across datasets. However, spurious attributes do not adhere to this assumption; they are contingent on the specific characteristics of the dataset. For instance, vehicle images in different datasets might be taken on a highway or in a parking lot, resulting in vastly different spurious attributes. This highlights the need for visual information from the dataset itself to accurately identify spurious attributes.

For \texttt{SAS}, the main concern still lies in efficiency. While the use of synthetic or pre-training images has been employed to address data scarcity in many recent works, such as SuS-X~\citep{udandarao2022sus} and Real-Prompt~\citep{parashar2024neglected}, these methods inevitably introduce additional computational overhead. The inference of Stable Diffusion~\citep{rombach2022high}, relative to its large data requirements, is not particularly fast, and retrieval requires finding top-k matches from a huge pre-training dataset~\citep{schuhmann2022laion}, both of which have efficiency bottlenecks. While selective optimization tricks can minimize computational burdens as much as possible, they come at the cost of accuracy.
\subsection{C.6 \quad More Related Works}
\label{subsec:C6}
{\textbf{Retrieval-Augmented Generation}. RAG is proposed essentially to address the insufficiency or lack of desired data. For example, \citet{long2022retrieval} improves long-tail recognition performance by retrieving text representations for tail classes. Similarly, \citet{parashar2024neglected} enhances VLMs' tail accuracy by identifying and retrieving high-frequency text synonyms corresponding to tail names from the training set. Furthermore, \citet{udandarao2022sus} mitigates data sparsity issues by retrieving external images through class names for data augmentation. Sharing motivations with previous work, we construct pseudo categories featuring spurious attributes through retrieval, thereby enhancing the model's robustness to these attributes. Nevertheless, beyond retrieval, we also explore data synthesis. In Section~\hyperref[subsec:B5]{B.5}, we compare the performance of our method using synthesized and retrieved data, empirically concluding that synthesized data yields greater accuracy gains. Compared to retrieval, synthesis can offer more tailored and precise scenarios and objects, which may be more suitable for our method given the diverse identified attributes.}

\section{D \quad Broader Societal Impacts}
\label{sec:D}
Our work has positive societal impacts. As illustrated in Fig.~5 of the main paper, VLMs may exhibit bias by associating harmful spurious attributes with target categories. For instance, when recognizing ${\rm street \ sign}$, VLMs often rely excessively on concepts like ${\rm street}$ and ${\rm road}$. This non-robust visual perception may lead to severe consequences in real-world applications, particularly in autonomous driving. The introduction of \texttt{SAP} can effectively identify such harmful attributes and even create a spurious attribute pool for specific applications, helping to determine situations where performance is compromised. Meanwhile, \texttt{SAS} provides an effective approach to suppress the influence of spurious attributes in VLMs, significantly enhancing the model's robustness against these attributes, including protected ones such as gender and race. Currently, we have not identified negative societal impacts of this work. However, due to objective factors, such as the availability of datasets and baselines' code, this will need to be further discussed in the future.
\section{E \quad Supplementary Results}
\label{sec:E}
\subsection{E.1 \quad Constructed Images}
\label{subsec:E1}
\begin{figure*}
    \centering
    \includegraphics[width=0.6\linewidth]{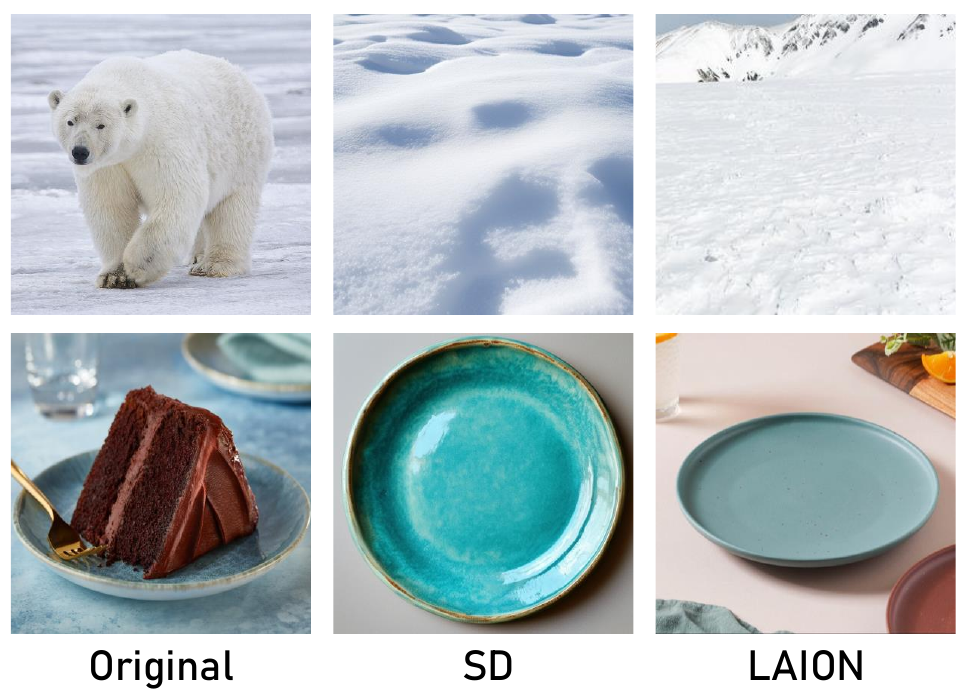}
    \caption{\textbf{More examples of generated and retrieved images with Stable Diffusion (SD) and LAION, respectively.
    } The category above is polar bear, with the primary spurious attribute being \textit{snow-covered ground}. The category below is chocolate cake, where one of the spurious attributes is the \textit{dinner plate}.}
\label{fig:more_examples}
\end{figure*}

In Fig.~\ref{fig:more_examples}, we provide more constructed images by \texttt{SAS}, with Stable Diffusion and retrieval from LAION-5B, respectively.
\subsection{E.2 \quad Spurious Attribute Statistics}
\label{subsec:E2}
Here, we present the spurious attribute statistics for the evaluated datasets. Specifically, we report the proportion of images containing one or more spurious attributes identified by \texttt{SAS} across 11 datasets, as shown in Table~\ref{tab:fpr}. The data reveals that for most datasets, over 50\% of images contain spurious attributes, highlighting the biased nature of these datasets and the consequent spurious correlations learned by VLMs.
\subsection{E.3 \quad Motivational Results}
\label{subsec:E3}
Given the enhanced generalization performance of VLMs before and after removing spurious attributes in Table 1 of the main paper, to further illustrate the impact of spurious attributes, here we present the improvement of the models on the counter group in Table~\ref{tab:preliminary_counter}. It can be observed that the accuracy of VLMs on the counter group shows a more significant improvement, up to 9\% on the unseen categories.
\subsection{E.4 \quad Numerical Main Results}
\label{subsec:E4}
Here we quantitatively demonstrate the main results as depicted in Fig.~3 of the main paper. Table~\ref{tab:base-to-new} and Table~\ref{tab:cross-dataset} present the numerical results of base-to-new generalization, cross-dataset transfer, and domain generalization, respectively.
\clearpage
\clearpage
\begin{table*}[t]
\footnotesize
\centering
\setlength{\tabcolsep}{0.8mm}
\renewcommand{\arraystretch}{1.6}
\begin{tabular}{cc}
\ChangeRT{1.2pt}
Dataset      & Images with spurious attributes (\%) \\ \hline
ImageNet     & 62.48                    \\
Caltech101   & 58.22                    \\
OxfordPets   & 73.54                    \\
StanfordCars & 69.92                    \\
Flowers102   & 63.50                    \\
Food101      & 54.59                    \\
FGVCAircraft & 47.97                    \\
SUN397       & 52.20                    \\
DTD          & 42.68                    \\
EuroSAT      & 47.90                    \\
UCF101       & 71.57                    \\ \ChangeRT{1.2pt}
\end{tabular}
\caption{\textbf{The proportion of images containing one or more spurious attributes of 11 datasets.}}
\label{tab:fpr}
\vspace{-8mm}
\end{table*}
\clearpage
\begin{table*}[t]
\footnotesize
\centering
\setlength{\tabcolsep}{0.8mm}
\renewcommand{\arraystretch}{1.6}
\begin{tabular}{lccccccccccccccc}
\ChangeRT{1.2pt}
\multicolumn{1}{c}{\multirow{2}{*}{Method}} & \multicolumn{3}{c}{FGVCAircraft} & \multicolumn{3}{c}{SUN397} & \multicolumn{3}{c}{Flowers102} & \multicolumn{3}{c}{DTD} & \multicolumn{3}{c}{Average} \\ \cline{2-16} 
\multicolumn{1}{c}{}                        & Base      & New      & SR      & Base      & New       & SR       & Base      & New      & SR      & Base   & New    & SR    & Base     & New     & SR     \\ \hline
CPL                                         & 29.25     & 24.80    & 5.43    & 63.47     & 54.74     & 6.61     & 76.98     & 60.77    & 5.71    & 68.93  & 43.14  & 5.13  & 59.66    & 45.86   & 5.72   \\
CPL - SA                                    & \textbf{32.42}     & \textbf{31.34}    &  \NA       & 64.81     & 60.35     &    \NA      & \textbf{78.35}     & \textbf{69.80}    &    \NA     & 71.26  & \textbf{52.65}  &     \NA  & 61.71    & \textbf{53.54}   &    \NA    \\ \hline
ArGue                                       & 27.24     & 21.92    & 5.13    & 65.23     & 57.44     & 6.45     & 72.26     & 59.32    & 6.69    & 70.77  & 40.12  & 5.97  & 58.87    & 44.70   & 6.06   \\
ArGue*                                      & 27.92     & 23.83    & 4.86    & 65.99     & 58.21     & 6.11     & 72.87     & 60.49    & 6.44    & 71.29  & 41.28  & 5.62  & 61.27    & 45.95   & 5.76   \\
ArGue - SA                                  & 30.71     & 29.20    &  \NA       & \textbf{67.90}     & \textbf{63.68}     &    \NA      & 74.68     & 66.45    &     \NA    & \textbf{73.70}  & 50.19  &   \NA    & \textbf{61.75}    & 52.38   &    \NA   \\ \ChangeRT{1.2pt}
\end{tabular}
\caption{\textbf{The results on the counter group in base-to-new generalization before and after removing spurious attributes (SA) from the pool.} We extract the counter group for both the base and new categories where spurious cues are removed. It can be observed that the accuracy of VLMs improves after removing spurious attributes in this context.}
\label{tab:preliminary_counter}
\vspace{-8mm}
\end{table*}
\clearpage
\begin{table*}[t]
\footnotesize
\centering
\setlength{\tabcolsep}{0.8mm}
\renewcommand{\arraystretch}{1.4}
\begin{adjustbox}{width=1.0\textwidth,center}
\begin{tabular}{lcccccccccccccccccccccc}
\ChangeRT{1.2pt}
\multicolumn{1}{c}{\multirow{2}{*}{Method}} & \multicolumn{2}{c}{ImageNet}                                            & \multicolumn{2}{c}{Caltech}                                    & \multicolumn{2}{c}{OxfordPets}                                 & \multicolumn{2}{c}{Cars}                                       & \multicolumn{2}{c}{Flowers}                                    & \multicolumn{2}{c}{Food}                                       & \multicolumn{2}{c}{Aircraft}                                   & \multicolumn{2}{c}{SUN}                               & \multicolumn{2}{c}{DTD}                                        & \multicolumn{2}{c}{EuroSAT}                           & \multicolumn{2}{c}{UCF}                                        \\ \cline{2-23} 
\multicolumn{1}{c}{}                        & Base                               & New                                & Base                      & New                                & Base                               & New                       & Base                      & New                                & Base                      & New                                & Base                      & New                                & Base                      & New                                & Base                      & New                       & Base                      & New                                & Base                      & New                       & Base                      & New                                \\ \hline
CLIP                                        & 72.43                              & 68.14                              & 96.84                     & 94.0                               & 91.17                              & 97.26                     & 63.37                     & 74.89                              & 72.08                     & 77.80                              & 90.10                     & 91.22                              & 27.19                     & 36.29                              & 69.36                     & 75.35                     & 53.24                     & 59.90                              & 56.48                     & 64.05                     & 70.53                     & 77.50                              \\ \hline
CoCoOp                                      & 75.98                              & 70.43                              & 97.96                     & 93.81                              & 95.20                              & 97.69                     & 70.49                     & 73.59                              & 94.87                     & 71.75                              & 90.70                     & 91.29                              & 33.41                     & 23.71                              & 79.74                     & 76.86                     & 77.01                     & 56.00                              & 87.49                     & 60.04                     & 82.33                     & 73.45                              \\
\quad + \texttt{SAS}                                       & 76.40                              & 71.38                              & 97.33                     & 94.62                              & 95.73                              & 97.80                     & 70.70                     & 74.96                              & 95.17                     & 72.99                              & 90.58                     & 92.12                              & 34.06                     & 28.30                              & 80.40                     & 79.67                     & 76.87                     & 57.73                              & 87.77                     & 63.61                     & 82.57                     & 75.66                              \\ \hline
KgCoOp                                      & 75.83                              & 69.96                              & 97.72                     & 94.39                              & 94.65                              & 97.76                     & 71.76                     & 75.04                              & 95.00                     & 74.73                              & 90.50                     & 91.70                              & 36.21                     & 33.55                              & 80.29                     & 76.53                     & 77.55                     & 54.99                              & 85.64                     & 64.34                     & 82.89                     & 76.67                              \\
\quad + \texttt{SAS}                                      & 75.78                              & 71.24                              & 97.96                     & 95.40                              & 95.43                              & 98.67                     & 71.46                     & 75.93                              & 95.51                     & 75.86                              & 90.69                     & 92.03                              & 36.93                     & 36.20                              & 80.80                     & 79.02                     & 77.89                     & 60.30                              & 85.78                     & 73.73                     & 83.38                     & 78.94                              \\ \hline
MaPLe                                       & 76.66                              & 70.54                              & 97.74                     & 94.36                              & 95.43                              & 97.76                     & 72.94                     & 74.00                              & 95.92                     & 72.46                              & 90.71                     & 92.05                              & 37.44                     & 35.61                              & 80.82                     & 78.70                     & 80.36                     & 59.18                              & 94.07                     & 73.23                     & 83.00                     & 78.66                              \\
\quad + \texttt{SAS}                                       & 76.69                              & 70.82                              & 97.92                     & 95.35                              & 95.88                              & 98.47                     & 73.16                     & 75.46                              & 95.93                     & 76.74                              & 91.41                     & 92.47                              & 37.87                     & 39.68                              & 81.30                     & 80.72                     & 80.78                     & 63.21                              & 94.38                     & 78.45                     & 82.89                     & 80.24                              \\ \hline
PromptSRC                                   & 77.60                              & 70.73                              & 98.10                     & 94.03                              & 95.33                              & 97.30                     & 78.27                     & 74.97                              & 98.07                     & 76.50                              & 90.67                     & 91.53                              & 42.73                     & 37.87                              & 82.67                     & 78.47                     & 83.37                     & 62.97                              & 92.90                     & 73.90                     & 87.10                     & 78.80                              \\
\quad + \texttt{SAS}                                       & 77.48                              & 71.48                              & 98.52                     & 95.20                              & 95.92                              & 98.50                     & 78.62                     & 75.24                              & 98.45                     & 79.11                              & 90.99                     & 92.43                              & 42.64                     & 40.16                              & 83.23                     & 80.80                     & 83.94                     & 63.46                              & 93.35                     & 76.68                     & \textbf{87.66}            & 81.55                              \\ \hline
LASP                                        & 76.25                              & 71.17                              & 98.17                     & 94.33                              & 95.73                              & 97.87                     & 75.23                     & 71.77                              & 97.17                     & 73.53                              & 91.20                     & 91.90                              & 38.05                     & 33.20                              & 80.70                     & 79.30                     & 81.10                     & 62.57                              & 95.00                     & 83.37                     & 85.53                     & 78.20                              \\
\quad + \texttt{SAS}                                       & 76.62                              & 72.45                              & 98.44                     & 95.27                              & 96.00                              & 98.58                     & 76.29                     & 72.85                              & 96.98                     & 75.30                              & 92.07                     & 92.65                              & 38.58                     & 33.97                              & 80.99                     & 80.42                     & 81.45                     & 64.11                              & \textbf{95.61}            & 83.69                     & 85.70                     & 79.19                              \\ \hline
TCP                                         & 77.27                              & 69.87                              & 98.23                     & 94.67                              & 94.67                              & 97.20                     & 80.80                     & 74.13                              & 97.73                     & 75.57                              & 90.57                     & 91.37                              & 41.97                     & 34.43                              & 82.63                     & 78.20                     & 82.77                     & 58.07                              & 91.63                     & 74.73                     & 87.13                     & 80.77                              \\
\quad + \texttt{SAS}                                       & 77.89                              & 70.53                              & 98.42                     & 95.40                              & 95.48                              & 98.23                     & 80.68                     & 75.80                              & 98.43                     & 75.48                              & 91.18                     & 92.70                              & 42.58                     & 35.10                              & \textbf{83.41}            & 79.27                     & 83.10                     & 60.67                              & 92.01                     & 75.34                     & 87.50                     & 81.62                              \\ \hline
CLIP-Adapter                                & 77.18                              & 70.25                              & 97.52                     & 93.48                              & 95.18                              & 96.43                     & 77.43                     & 72.64                              & 96.83                     & 71.75                              & 90.98                     & 90.55                              & 41.89                     & 33.10                              & 80.73                     & 77.98                     & 82.17                     & 58.72                              & 93.34                     & 71.84                     & 86.49                     & 77.38                              \\
\quad + \texttt{SAS}                                       & 77.09                              & 71.98                              & 98.10                     & 95.55                              & 95.72                              & 97.70                     & 77.63                     & 75.46                              & 97.59                     & 73.13                              & 90.74                     & 91.38                              & 41.88                     & 36.50                              & 81.42                     & 80.15                     & 82.84                     & 61.27                              & 93.99                     & 73.87                     & 86.87                     & 78.51                              \\ \hline
Tip-Adapter                                 & 78.04                              & 71.96                              & 98.68                     & 94.17                              & 96.21                              & 98.57                     & 80.79                     & 73.66                              & 98.73                     & 74.36                              & 92.62                     & 91.01                              & \textbf{43.34}            & 35.73                              & 81.77                     & 79.27                     & 84.58                     & 59.91                              & 94.82                     & 74.81                     & 86.78                     & 78.94                              \\
\quad + \texttt{SAS}                                       & 77.89                              & 72.58                              & \textbf{98.89}            & 95.72                              & 96.65                              & 98.43                     & \textbf{80.84}            & 75.42                              & \textbf{98.81}            & 76.73                              & \textbf{92.84}            & 92.81                              & 43.27                     & 38.31                              & 82.36                     & 80.13                     & \textbf{84.77}            & 63.38                              & 95.51                     & 77.54                     & 87.42                     & 79.77                              \\ \hline
ArGue                                       & 76.95                              & 71.86                              & 98.63                     & 94.70                              & 96.23                              & 98.59                     & 75.06                     & 74.18                              & 98.62                     & 77.96                              & 91.42                     & 92.40                              & 41.29                     & 38.80                              & 81.89                     & 80.48                     & 80.33                     & 67.03                              & 95.10                     & 90.68                     & 86.00                     & 79.43                              \\
\quad + \texttt{SAP}                                       & 77.32                              & 72.04                              & 98.57                     & 95.12                              & 96.34                              & 98.86                     & 75.72                     & 74.90                              & 98.66                     & 78.78                              & 91.54                     & 92.63                              & 41.86                     & 39.65                              & 82.43                     & 81.78                     & 80.87                     & 68.36                              & 95.46                     & 91.51                     & 86.59                     & 80.28                              \\
\quad + \texttt{SAS}                                       & 77.59                              & 72.36                              & 98.69                     & 95.88                              & 96.52                              & 98.75                     & 76.24                     & 75.51                              & 98.74                     & 79.65                              & 91.81                     & 93.42                              & 42.39                     & 40.84                              & 82.71                     & 82.21            & 81.35                     & \textbf{69.73}                              & 95.41                     & \textbf{92.47}            & 87.05                     & 81.73                              \\ \hline
MAP                                         & 76.60                              & 70.60                              & 98.30                     & 93.80                              & 95.43                              & 96.90                     & 76.70                     & 73.73                              & 97.57                     & 75.23                              & 90.30                     & 89.30                              & 41.63                     & 36.43                              & 82.33                     & 76.30                     & 82.63                     & 66.23                              & 92.13                     & 76.10                     & 86.67                     & 78.77                              \\
\quad + \texttt{SAP}                                       & 76.73                              & 71.17                              & 98.21                     & 94.32                              & 95.79                              & 98.09                     & 77.34                     & 74.10                              & 97.85                     & 77.55                              & 90.60                     & 90.76                              & 42.05                     & 37.72                              & 82.15                     & 78.05                     & 82.77                     & 67.61                              & 92.53                     & 77.22                     & 87.09                     & 79.42                              \\
\quad + \texttt{SAS}                                       & 76.79                              & 72.25                              & 98.53                     & 94.52                              & 96.19                              & 98.83                     & 77.70                     & 75.80                              & 97.79                     & 79.99                              & 90.89                     & 91.63                              & 42.33                     & 39.32                              & 82.74                     & 78.47                     & 82.87                     & 68.44                              & 93.30                     & 78.21                     & 87.44                     & 80.85                              \\ \hline
CPL                                         & 78.74                              & 72.03                              & 98.35                     & 95.13                              & 95.86                              & 98.21                     & 79.31                     & 76.65                              & 98.07                     & 80.43                              & 91.92                     & 93.87                              & 42.27                     & 38.85                              & 81.88                     & 79.65                     & 80.92                     & 62.27                              & 94.18                     & 81.05                     & 86.73                     & 80.17                              \\
\quad + \texttt{SAP}                                       & 78.76                              & 72.64                              & 98.67                     & 95.72                              & 96.31                              & \textbf{98.87}            & 79.24                     & 78.12                              & 98.37                     & 82.51                              & 92.19                     & 94.63                              & 42.15                     & 40.92                              & 82.16                     & 81.62                     & 82.21                     & 65.72                              & 94.47                     & 83.55                     & 86.51                     & 80.91                              \\
\quad + \texttt{SAS}                                       & \multicolumn{1}{l}{\textbf{78.82}} & \multicolumn{1}{l}{\textbf{73.49}} & \multicolumn{1}{l}{98.59} & \multicolumn{1}{l}{\textbf{95.98}} & \multicolumn{1}{l}{\textbf{96.76}} & \multicolumn{1}{l}{98.82} & \multicolumn{1}{l}{79.77} & \multicolumn{1}{l}{\textbf{80.35}} & \multicolumn{1}{l}{98.71} & \multicolumn{1}{l}{\textbf{83.46}} & \multicolumn{1}{l}{92.26} & \multicolumn{1}{l}{\textbf{95.45}} & \multicolumn{1}{l}{42.61} & \multicolumn{1}{l}{\textbf{41.72}} & \multicolumn{1}{l}{82.11} & \multicolumn{1}{l}{\textbf{83.17}} & \multicolumn{1}{l}{83.00} & \multicolumn{1}{l}{67.89} & \multicolumn{1}{l}{94.75} & \multicolumn{1}{l}{87.07} & \multicolumn{1}{l}{87.21} & \multicolumn{1}{l}{\textbf{82.22}} \\ \ChangeRT{1.2pt}
\end{tabular}
\end{adjustbox}
\caption{\textbf{The numerical results on base-to-new generalization.}}
\label{tab:base-to-new}
\end{table*}
\clearpage
\begin{table*}[t]
\footnotesize
\centering
\setlength{\tabcolsep}{0.8mm}
\renewcommand{\arraystretch}{1.4}
\begin{adjustbox}{width=1.0\textwidth,center}
\begin{tabular}{llclcccccccccclcccc}
\ChangeRT{1.2pt}
\multirow{3}{*}{} &  & \textbf{Source}      &  & \multicolumn{10}{c}{\textbf{Cross-dataset Transfer Target}}                                                                                                                                                                         &  & \multicolumn{4}{c}{\textbf{Domain Generalization Target}}                                 \\ \cline{3-3} \cline{5-14} \cline{16-19} 
                  &  & \multicolumn{1}{l}{} &  & \multicolumn{1}{l}{} & \multicolumn{1}{l}{} & \multicolumn{1}{l}{} & \multicolumn{1}{l}{} & \multicolumn{1}{l}{} & \multicolumn{1}{l}{} & \multicolumn{1}{l}{} & \multicolumn{1}{l}{} & \multicolumn{1}{l}{} & \multicolumn{1}{l}{} &  & \multicolumn{1}{l}{} & \multicolumn{1}{l}{} & \multicolumn{1}{l}{} & \multicolumn{1}{l}{} \\
                  &  & \rotbox{ImageNet}             &  & \rotbox{Caltech}              & \rotbox{Pets}                 & \rotbox{Cars}                 & \rotbox{Flowers}              & \rotbox{Food}                 & \rotbox{Aircraft}             & \rotbox{SUN}                  & \rotbox{DTD}                  & \rotbox{EuroSAT}              & \rotbox{UCF}                  &  & \rotbox{ImageNet-V}           & \rotbox{ImageNet-S}      & \rotbox{ImageNet-A}           & \rotbox{ImageNet-R}           \\ \hline
CLIP              &  & 66.54                &  & 94.62                & 90.41                & 64.69                & 70.30                & 85.63                & 23.73                & 66.12                & 44.84                & 47.50                & 67.42                &  & 63.20                & 48.35                & 49.32                & 76.57                \\ \hline
CoCoOp            &  & 71.02                &  & 94.43                & 90.14                & 65.32                & 71.88                & 86.06                & 22.94                & 67.36                & 45.73                & 45.37                & 68.21                &  & 64.07                & 48.75                & 50.63                & 76.18                \\
\quad +\texttt{SAS}              &  & 71.35                &  & 95.59                & 90.84                & 66.76                & 72.47                & 86.34                & 23.81                & 68.99                & 47.94                & 46.90                & 70.84                &  & 64.97                & 49.56                & 51.61                & 77.31                \\ \hline
KgCoOp            &  & 70.66                &  & 93.92                & 89.83                & 65.41                & 70.01                & 86.36                & 22.51                & 66.16                & 46.35                & 46.04                & 68.50                &  & 64.10                & 48.97                & 50.69                & 76.70                \\
\quad +\texttt{SAS}              &  & 70.90                &  & 94.33                & 89.68                & 67.82                & 71.13                & 88.91                & 24.60                & 67.47                & 47.72                & 48.22                & 68.52                &  & 64.53                & 49.72                & 51.70                & 77.22                \\ \hline
MaPLe             &  & 70.72                &  & 93.53                & 90.49                & 65.57                & 72.23                & 86.20                & 24.74                & 67.01                & 46.49                & 48.06                & 68.69                &  & 64.07                & 49.15                & 50.90                & 76.98                \\
\quad +\texttt{SAS}              &  & 71.21                &  & 93.61                & 91.76                & 67.53                & 73.60                & 87.58                & 24.54                & 67.69                & 47.98                & 48.17                & 71.96                &  & 63.98                & 50.74                & 51.57                & 77.25                \\ \hline
PromptSRC         &  & 71.27                &  & 93.60                & 90.25                & 65.70                & 70.25                & 86.15                & 23.90                & 67.10                & 46.87                & 45.50                & 68.75                &  & 64.35                & 49.55                & 50.90                & 77.80                \\
\quad +\texttt{SAS}              &  & 71.53                &  & 93.25                & 92.60                & 66.44                & 70.13                & 88.19                & 25.05                & 67.87                & 47.22                & 45.50                & 68.99                &  & 64.07                & 50.40                & 51.52                & \textbf{78.98}       \\ \hline
LASP              &  & 71.34                &  & 93.65                & 91.83                & 67.29                & 70.82                & 88.54                & 28.60                & 65.75                & 54.83                & 43.65                & 69.23                &  & 64.04                & 47.93                & 49.11                & 75.36                \\
\quad +\texttt{SAS}              &  & 71.62                &  & 94.62                & 92.98                & 68.89                & 71.18                & 89.89                & 29.68                & 68.47                & 55.74                & 45.80                & 71.63                &  & 65.24                & 47.91                & 50.80                & 77.08                \\ \hline
TCP               &  & 71.40                &  & 93.97                & 91.25                & 64.69                & 71.21                & 86.69                & 23.45                & 67.15                & 44.35                & 51.45                & 68.73                &  & 64.60                & 49.50                & 51.20                & 76.73                \\
\quad +\texttt{SAS}              &  & 71.73                &  & 94.73                & 92.60                & 66.54                & 71.44                & 87.81                & 24.80                & 68.94                & 45.15                & 52.93                & 70.31                &  & 65.62                & 50.79                & \textbf{52.94}       & 78.82                \\ \hline
CLIP-Adapter      &  & 72.35                &  & 93.06                & 90.76                & 63.17                & 69.23                & 85.13                & 20.54                & 65.57                & 43.27                & 49.64                & 66.33                &  & 62.91                & 49.15                & 51.74                & 76.81                \\
\quad +\texttt{SAS}              &  & 72.53                &  & 93.12                & 91.72                & 66.65                & 69.18                & 88.10                & 22.27                & 66.60                & 45.69                & 50.38                & 69.80                &  & 64.50                & 49.70                & 52.39                & 77.75                \\ \hline
Tip-Adapter       &  & 72.53                &  & 95.71                & 93.12                & 66.61                & 68.83                & 89.22                & 23.63                & 68.32                & 47.31                & 53.40                & 68.15                &  & 63.30                & 49.26                & 50.18                & 76.70                \\
\quad +\texttt{SAS}              &  & 72.81                &  & 95.49                & \textbf{94.88}       & 67.80                & 68.46                & 91.77                & 25.00                & 69.46                & 49.55                & \textbf{54.33}       & 68.94                &  & 64.21                & 50.34                & 50.89                & 77.93                \\ \hline
ArGue             &  & 71.84                &  & 94.20                & 92.66                & 70.70                & 71.29                & 91.64                & 28.28                & 70.51                & 55.37                & 45.76                & 71.97                &  & 65.02                & 49.25                & 51.47                & 76.96                \\
\quad +\texttt{SAP}              &  & 72.14                &  & 95.74                & 93.75                & 71.80                & 72.48                & 91.87                & 28.53                & 70.88                & 56.54                & 46.86                & 72.96                &  & 65.47                & 49.94                & 52.48                & 77.38                \\
\quad +\texttt{SAS}              &  & 72.28                &  & 95.67                & 94.29                & \textbf{72.72}       & 74.63                & \textbf{92.53}       & 29.10                & \textbf{71.96}       & \textbf{57.40}       & 48.22                & \textbf{73.82}       &  & 66.12                & 49.90                & 52.85                & 77.90                \\ \hline
MAP               &  & 71.60                &  & 93.93                & 90.80                & 63.00                & 68.40                & 86.07                & 24.87                & 68.10                & 51.87                & 42.63                & 68.73                &  & 64.47                & 49.07                & 51.07                & 77.37                \\
\quad +\texttt{SAP}              &  & 71.93                &  & 95.40                & 92.63                & 64.50                & 68.13                & 87.18                & 26.80                & 69.99                & 51.35                & 44.10                & 70.50                &  & 65.06                & 49.88                & 51.64                & 77.34                \\
\quad +\texttt{SAS}              &  & 72.21                &  & 95.82                & 93.73                & 66.69                & 68.46                & 88.11                & 28.62                & 70.29                & 51.91                & 45.73                & 71.59                &  & 66.14                & 50.78                & 52.19                & 77.70                \\ \hline
CPL               &  & 73.53                &  & 95.52                & 91.64                & 66.17                & 73.35                & 87.68                & 27.36                & 68.24                & 48.96                & 51.25                & 70.52                &  & 65.24                & 50.84                & 52.10                & 76.76                \\
\quad +\texttt{SAP}              &  & 73.75                &  & \textbf{95.83}       & 92.92                & 66.69                & 74.32                & 88.33                & 29.58                & 69.64                & 49.81                & 52.72                & 71.35                &  & \textbf{66.45}       & 51.93                & 52.61                & 77.74                \\
\quad +\texttt{SAS}              &  & \textbf{73.94}       &  & 95.74                & 93.67                & 67.22                & \textbf{75.67}       & 89.49                & \textbf{30.55}       & 70.26                & 49.91                & 54.29                & 72.48                &  & 66.38                & \textbf{52.95}       & 52.81                & 78.31                \\ \ChangeRT{1.2pt}
\end{tabular}
\end{adjustbox}
\caption{\textbf{The numerical results on cross-dataset transfer and domain generalization.}}
\label{tab:cross-dataset}
\end{table*}
\end{document}